\documentclass[12pt]{article}
\usepackage[T1]{fontenc}
\usepackage{amssymb}
\usepackage{amsmath}
\usepackage{bm}
\usepackage{graphicx}
\usepackage{placeins}
\usepackage{float}

\usepackage{xcolor}
\usepackage{hyperref}
\hypersetup{
    colorlinks=true,
    linkcolor=black,
    citecolor=black,
    urlcolor=blue
}
\usepackage[acronym]{glossaries}
\makenoidxglossaries
\usepackage[numbers,sort&compress]{natbib}

\usepackage{tabularx}
\usepackage{booktabs}
\usepackage{multirow}
\usepackage{siunitx}
\usepackage{natbib}
\usepackage{soul}
\usepackage{multicol}
\usepackage{lmodern}
\usepackage{comment}
\usepackage{xr}
\usepackage{dblfloatfix}
\usepackage[font=small,labelfont=bf]{caption}

\usepackage[margin=1in]{geometry}
\newacronym{rl}{RL}{Reinforcement Learning}
\newacronym{rnn}{RNN}{Recurrent Neural Network}
\newacronym{pca}{PCA}{Principal Component Analysis}
\newacronym{mcmc}{MCMC}{Markov chain Monte Carlo}
\newacronym{mve}{MVE}{Mean–Variance Estimation}
\newacronym{mpo}{MPO}{Multi-parameter Optimization}
\newacronym{ad}{AD}{Applicability Domain}
\newacronym{cp}{CP}{Conformal prediction}
\newacronym{ms}{MS}{Model System}
\newacronym{rf}{RF}{Random Forest}
\newacronym{ecfp}{ECFP}{Extended-connectivity Fingerprints}
\newacronym{icp}{ICP}{Inductive Conformal Predictor}
\newacronym{tpsa}{TPSA}{Topological Polar Surface Area}
\newacronym{smi}{SMILES}{Simplified Molecular Input Line Entry System}
\newacronym{mw}{MW}{Molecular Weight}
\newacronym{qsar}{QSAR}{Quantitative Structure-Activity Relationship}
\newacronym{drd2}{DRD2}{Dopamine Receptor Type 2}
\newacronym{egfr}{EGFR}{Epidermal Growth Factor Receptor}
\newacronym{si}{SI}{Supplementary Information}
\newacronym{sm}{SM}{Score Modulation}
\newacronym{lm}{LM}{Loss Modulation}

\newcommand{\addMolAI}{Molecular AI, Discovery Sciences, BioPharmaceuticals R\&D}
\newcommand{\addAstra}{AstraZeneca AB}
\newcommand{\addAddress}{Gothenburg, Sweden}

\title{Uncertainty-aware reinforcement learning for chemical language models}

\author{
Borja Medina\\
\addMolAI\\
\addAstra\\
\addAddress\\
\texttt{borja.medinadelasheras@astrazeneca.com}
\and
Jon Paul Janet\\
\addMolAI\\
\addAstra\\
\addAddress
}

\date{}

\begin{document}

\maketitle

\abstract{
    Reinforcement Learning (RL) has become a powerful paradigm for \textit{de novo} molecular design, enabling Chemical Language Models (CLMs) to navigate and explore the chemical space while optimizing specific desired properties. However, the existing RL frameworks treat all scoring functions as deterministic oracles, neglecting the inherent uncertainty attached to the predictions of the different molecular properties. This can lead to the exploration of highly-uncertain regions of the chemical space, focusing on the generation of highly scored molecules which are poorly supported by the training data. This can destabilize the optimization process, yielding predictions that are far from their true values.
    
    We propose and compare two complementary ways of incorporating predictive uncertainty into RL. In the first one, uncertainty is treated as an additional optimization objective and incorporated along with the rest of the scoring functions, allowing the policy to trade off exploitation against reliability. Secondly, uncertainty is used to modulate policy updates, reducing the influence of molecules whose properties lie far outside the scoring function confidence domain.

    Both approaches were evaluated across three different settings: (i) a controlled model system, in which the prediction error is modeled as a Gaussian distribution, with a variance proportional to the distance to the training data; and two real-world tasks, making use of (ii) ChemProp models and (iii) a Conformal Prediction wrapper applied to a Random forest classifier.
    
    We show that uncertainty-aware RL enables CLMs to explore chemical space more robustly by favoring lower-uncertainty regions. This leads to more reliable hit discovery without compromising molecular score, increasing the true hit rate by 0.25 (from 0.5 to 0.75), and nearly doubling the total number of true hits.
}

\noindent\textbf{Keywords:} drug design, generative AI, Reinforcement Learning, uncertainty

\maketitle

\clearpage

\section{Introduction}\label{Intro}

Designing novel small molecules with desired physicochemical and biological properties is a central challenge in drug discovery and molecular engineering, with broad implications for human health \cite{Sadybekov2023_ComputationalApproachesDrugDiscovery, Ozelik2025_DenovoDrugDesignReview}, environmental sustainability \cite{SandovalPauker2023_ComputationalChemistryEnviroment,Leonard2021_AISustainableChemistry}, and materials science \cite{Achar2024_MLMaterialScience,Duan2022_MLMaterialSciences}. 

The chemical space of small molecules is immense, estimated to contain more than $10^{60}$ different small organic molecules \cite{Reymond2012_EnumerationChemicalSpace}. Consequently, exhaustive experimental exploration becomes infeasible. As a result, computational methods have become essential tools for navigating the chemical space and accelerating drug discovery.  

Among the different computational methods, deep generative modeling has enabled data-driven exploration of the chemical space by learning the underlying distribution over molecular representations, allowing the sampling of novel molecular candidates \cite{Ozelik2025_DenovoDrugDesignReview}. However, although these models are extremely effective in generating novel molecules, they do not inherently steer the exploration toward desired properties. 

With demonstrated success across a range of domains such as games \cite{Silver2018_RLinGames,Berner2019_RLinDota2}, robotics \cite{Singh2021_RLinRobotics}, autonomous driving \cite{Talpaert2019_RLinDriving,Kiran2022_RLinDriving}, \gls{rl} has become a widely adopted framework for guiding optimization in drug discovery \cite{Korshunova2022_GenerativeRLBioactiveCompounds,Zhou2019_OptimizationDeepRL,Stahl2019_DeepRLOptimization, Olivecrona2017_MolecularDenovoDeepRL, Popova2018_RLinDrugDesign}. 

In \gls{rl}-based molecular generation, a generative deep learning model is treated as a policy that proposes new molecules, which are then evaluated, making use of predictive models \cite{sutton1998_RL, Olivecrona2017_MolecularDenovoDeepRL, Popova2018_RLinDrugDesign}. This paradigm is shared among several frameworks, including REINVENT \cite{Loeffler2024_REINVENT4}, ACEGEN \cite{Bou2024_ACEGEN}, Saturn \cite{Guo2026_Saturn}, among others. In this work we make use of REINVENT, which employs a likelihood-based policy update to steer generation towards high-scoring regions of the molecular space.

\begin{figure}[ht]
    \centering
    \includegraphics[page=1,width=0.8\linewidth,trim=3.8cm 3cm 5cm 3.8cm,clip]{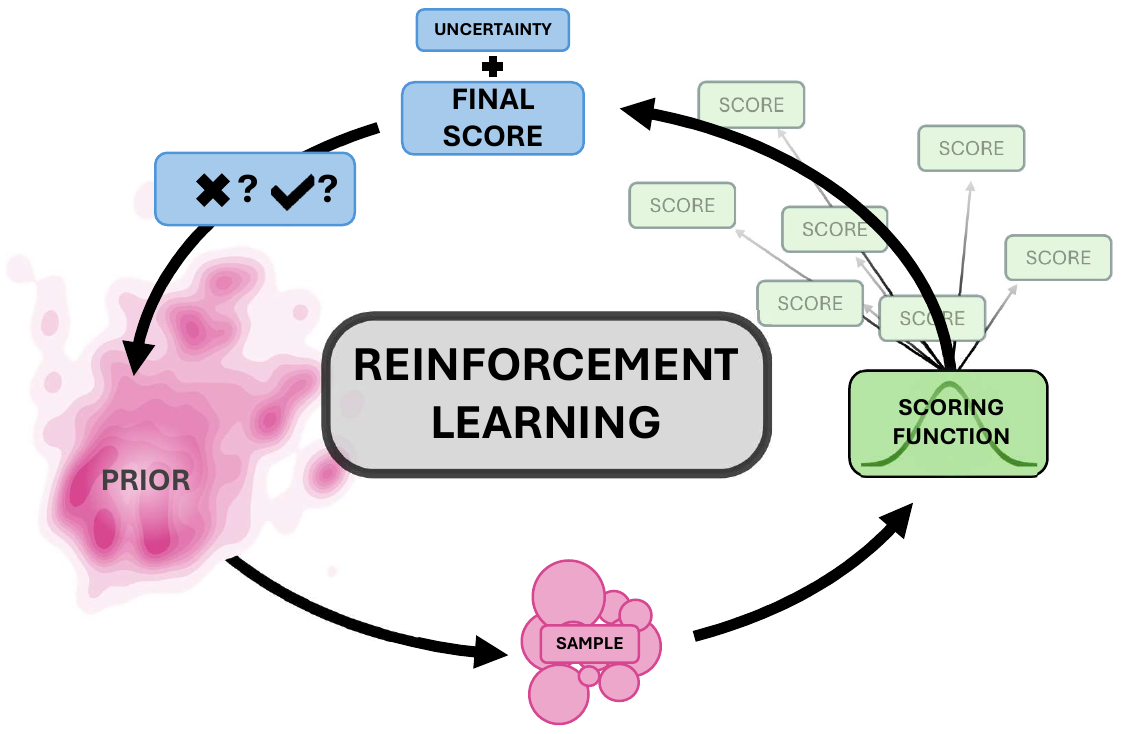}
    \caption{\gls{rl} setup in REINVENT4. The framework consists of an RNN prior that samples molecules from a learned chemical distribution. Generated molecules are evaluated using multiple scoring functions, which provide scalar predictions; these predictions can be interpreted as samples from an underlying distribution. The individual scalar scores are aggregated into a single final score used for the \gls{rl} optimization process. Although scoring functions are not deterministic oracles and exhibit predictive uncertainty, this uncertainty is not explicitly incorporated into the REINVENT4 optimization procedure. In this paper, we propose different strategies for incorporating this uncertainty into the optimization process.}
    \label{fig:rl}
\end{figure}

A common limitation of current molecular design frameworks is the treatment of the scoring functions as deterministic oracles \cite{LockWood2022_ReviewUncertaintyDeepRL}. In practice, these scoring functions are subject to multiple sources of uncertainty. The predictive models embedded in these scoring functions are trained on limited and potentially biased datasets, and therefore their outputs are inherently uncertain. Moreover, uncertainty is not restricted to data-driven components; even non-data-driven scoring terms may be uncertain due to approximations in the scoring formulation, sensitivity to parameter choices, numerical errors, or variability across alternative scoring implementations. Broadly, these uncertainties can be decomposed in terms of aleatoric uncertainty, arising from noise and stochasticity in the data, and epistemic uncertainty, which reflects uncertainty in the model parameters and limited coverage of the training domain \cite{Kendall2017_Uncertainties, Mervin2021_UQDrugDesign}. 

Several complementary approaches have been proposed to quantify uncertainty in molecular property prediction. These include Bayesian methods \cite{Wagenmakers2008_BayesianFrequentist, willink2012_BayesianUncertainty, Tran2020_UQinMAterialScience, Mervin2021_UQDrugDesign}, and Frequentist methods \cite{Wagenmakers2008_BayesianFrequentist, Mervin2021_UQDrugDesign, Kompa2021_FrequentistUncertainty, dorigatti2023_FrequentistUncertainty}. In addition, although they are not uncertainty measures in the strict probabilistic sense, \gls{cp} and other \gls{ad}-based methods are widely used to assess prediction reliability. 

Bayesian approaches quantify uncertainty through posterior inference over model parameters or functions. Common implementations include variational inference methods, which approximate the posterior distribution, and \gls{mcmc} techniques~\cite{Mervin2021_UQDrugDesign}. In contrast, Frequentist approaches do not rely on explicit posterior distributions, instead, they estimate predictive uncertainty from observable variability in predictions. For example, deep ensembles, which estimate uncertainty from the variability across independently trained models \cite{Mervin2021_UQDrugDesign}. Methods such as \gls{mve} extend these ideas to predict both the mean and the variance of the target distribution, capturing the aleatoric uncertainty.

\glspl{ad} do not constitute an uncertainty measure in the strict sense. Rather, they have traditionally been defined as the region of chemical space in which a model is expected to make reliable predictions \cite{Tropsha2003_preAD, Dimitrov2005_ADdefinition, Eriksson2003_ADinQSAR, Gadaleta2016_ADinQSAR, Sahigara2012_ADinQSAR}. It is typically assessed using similarity or distance measures between query molecules and the training data in a chosen feature space. While molecules closer to the training distribution are often assumed to yield more reliable predictions, in reality, \gls{ad} does not provide a guarantee of predictive accuracy. Instead, it serves as a proxy for whether a query lies in an interpolative region relative to the training data.  Despite lacking an unambiguous mathematical formulation, \gls{ad} methods are widely used in \gls{qsar} modeling to assess prediction reliability and guide chemical space exploration \cite{Schultz2025_ApplicabilityDomain, Eriksson2003_ADinQSAR, Gadaleta2016_ADinQSAR, Sahigara2012_ADinQSAR} and can be interpreted as a proxy for the presence of epistemic uncertainty.

\gls{cp} is also widely used for uncertainty quantification. It provides a distribution-free framework that can be applied as a wrapper around any predictive model. Instead of explicitly modeling uncertainty, \gls{cp} constructs calibrated prediction intervals or sets based on observed residuals \cite{Angelopoulos2023_CP}. By design, \gls{cp} does not explicitly distinguish between aleatoric and epistemic uncertainty; rather, it reflects the overall predictive uncertainty present in the calibration data.

Nevertheless, despite their widespread use, many molecular design frameworks, including REINVENT4 \cite{Loeffler2024_REINVENT4}, typically treat scoring functions as deterministic, ignoring the uncertainty associated with the predictions \cite{lee2023_exploringchemicalspacescorebased,Alshehri2025_UncertaintyGuidanceDifussion}, and therefore, assuming that all generated molecules lie within a reliable prediction region. This idea may not hold in practice, especially when exploring regions of chemical space that are far from the training distribution. Generative models have been shown to have the ability to explore the out-of-distribution regions of the training chemical space \cite{Zhang2021_ChemicalSpaceCoverage, Renz2019_failureModes}, producing molecules for which the scoring functions have little predictive validity. Therefore, not taking into account the uncertainty associated with the scoring functions can lead to over-exploitation of unreliable high-scoring regions \cite{lee2023_exploringchemicalspacescorebased, Yoshizawa2025_DyRAMO}.  In other words, the agent could shift exploration into undesirable regions of the molecular space, where predictions are highly uncertain, producing unrealistic predictions or chemically implausible molecules, destabilizing the optimization process.

While predictive uncertainty has been widely studied in \gls{qsar} modeling and molecular property prediction \cite{Eriksson2003_ADinQSAR,Sahlin2013_UQinQSAR, Hirschfeld2020_UQinNN, Frombgen2026_UQReview, Rasmussen2023_UQforChemicalData}, it has received limited attention in the context of \gls{rl}-based molecular generation \cite{Maranas1997_MolecularDesignUncertainty, Yoshizawa2025_DyRAMO}. In this work, we address this gap by introducing uncertainty-aware \gls{rl} for \textit{de novo} molecular design within the REINVENT4 framework. We propose two different strategies for incorporating predictive uncertainty into the \gls{rl} loop. The first one treats uncertainty as an explicit optimization objective, while the second one uses uncertainty to modulate the magnitude of policy updates. 

We evaluate these strategies using (i) a synthetic Model System designed to resemble an \gls{ad} setting, where uncertainty is informative by design. This setup acts as a controlled perfect model scenario in which the magnitude of prediction error can be explicitly adjusted. In addition, we assess performance in two real-data scenarios: (ii) ChemProp models trained with \gls{mve} to capture aleatoric uncertainty, and (iii) \gls{rf} models combined with conformal prediction to provide calibrated prediction sets. 

We show that both strategies improve the stability and reliability of molecular optimization in the different scenarios, while preserving the ability to explore high-scoring compounds.


\section{Methods}\label{Methods}

\subsection{Datasets}\label{MethodsDatasets}

\begin{table*}[h]
    \centering
    \caption{
        Overview of the training datasets and the corresponding mean and standard deviation of their $pIC_{50}$ values. N/A means that $IC_{50}$ values are not used during the study, only \gls{smi} are used as model input. Additionally, we also report the mean \gls{mw} and the mean \gls{tpsa}
    }
    \label{tabDatasets}
    \begin{tabularx}{0.8\linewidth}{@{}X >{\raggedleft\arraybackslash}X >{\raggedleft\arraybackslash}X >{\raggedleft\arraybackslash}X>{\raggedleft\arraybackslash}X>{\raggedleft\arraybackslash}X >{\raggedleft\arraybackslash}X@{}}
        \textbf{Dataset} & \textbf{\textbf{N$_{\text{molecules}}$}} & \textbf{Mean $pIC_{50}$} & \textbf{std $pIC_{50}$} & Mean MW & Mean TPSA\\
        \toprule
        $\gls{drd2}_{\text{training}}$    & $2\,200$ &  $N/A$  & $N/A$ & $428.82$ & $70.50$\\
        $\gls{egfr}_{\text{training}}$    &  $2\,200$ &  $N/A$   &  $N/A$ & $483.72$ & $110.06$\\
        $\gls{egfr}_{\text{small}}$    &  $800$ &  $6.976$   &  $1.247$ & $479.84$ & $108.04$\\        
        $\gls{egfr}_{\text{full}}$     & $36\,220$  &   $6.947$   &   $1.275$  & $481.28$ & $109.21$ \\
        
\bottomrule
\end{tabularx}
\end{table*}

The training datasets were constructed using molecules retrieved from the \textit{ChEMBL} database (version 34). We retrieved all compounds with reported $pIC_{50}$ values against the \gls{drd2} and the \gls{egfr}, creating two different datasets. For each dataset, basic cleaning was performed, including canonicalization and deduplication of the \gls{smi}. For more information about the cleaning procedure see \textit{\gls{si} Section~1}.

The complete set of cleaned \gls{egfr} compounds with reported $pIC_{50}$ values retrieved from ChEMBL is referred to as $\gls{egfr}_{\text{full}}$. From this dataset 36\,220 (80\%) active molecules were reserved for training, 4\,527 (10\%) for validation and 4\,528 (10\%) for testing purposes. While keeping same validation and test sets, the training set was randomly subsampled down to 800 molecules creating the  $\gls{egfr}_{\text{small}}$. These two \gls{egfr} datasets were used for training the ChemProp models.

Additionally, from each target-specific ChEMBL dataset, $2\,200$ molecules were randomly selected to generate two \gls{smi} training datasets, hereafter referred to as $\gls{drd2}_{\text{training}}$ and $\gls{egfr}_{\text{training}}$. They were used as the reference datasets to construct the \gls{ad}.

\subsection{REINVENT4}\label{MethodsREINVENT4}
REINVENT4 is an open-source framework for molecular design that includes a \textit{de novo} generation model (reinvent), based on a \gls{rnn} trained on data derived from ChEMBL version 22 \cite{Loeffler2024_REINVENT4, Olivecrona2017_MolecularDenovoDeepRL, Blaschke2020_reinvent2}. This \gls{rnn} prior is capable of generating valid \gls{smi} across broad regions of the molecular space. 

\gls{rl} could be applied to this prior, to steer the generation process toward specific regions of the space with desirable molecular properties, quantified by a \gls{mpo} score. This exploration is controlled by the augmented likelihood ($\log P_{\text{aug}}$), which represents a balance between the prior likelihood and the \gls{mpo} scoring function and acts as a reward signal for the agent. At initialization, the agent is initialized with the same parameters as the prior.
\begin{equation}
\log P_{\text{aug}}(x) = \log P_{\text{prior}}(x) + \text{MPO}(x)
\label{augmentedLikelihood}
\end{equation}

where $x$ denotes a SMILES string representing a molecule.

This way the prior likelihood $\log P_{\text{prior}}$ acts as a regularization term, ensuring that generated molecules remain chemically plausible while optimizing for desired properties.

The \gls{mpo} score in REINVENT is defined as a weighted combination of individual scoring components. Depending on the formulation, this score can be expressed as a geometric  or arithmetic mean of the transformed component scores. In this work we made use of the geometric mean:

\begin{equation}
\mathrm{MPO}(x) = \prod_{i=1}^{n} \theta_s \left(S_i(x)\right)^{\frac{w_i}{\sum_{l=1}^{n}w_l}}
\label{MPO}
\end{equation}

where $S_i(x)$ denotes the $i$-th scoring component, $\theta_s$ is a transformation function mapping each score into the range $[0,1]$, and $w_i$ are the corresponding weights,  that determine the relative importance of each scoring component (in our case $w_i=1,\forall i $).

At each epoch, the agent is updated by minimizing the following loss function making use of the wDAP learning strategy \cite{Fialkov2021_wDAP}.

\begin{equation}
\mathcal{L}(X)
= \frac{1}{N} \sum_{x \in X}
\left( \log P_{\mathrm{aug}}(x) - \log P_{\mathrm{agent}}(x) \right)^2
\label{lossDefinition}
\end{equation}

More information on the different scoring functions can be found in \textit{\gls{si} Section~2}.

\subsubsection{Probabilistic scoring functions} \label{methods:ProbScoringFucntions}
\begin{figure}[!htbp]
    \centering
    \includegraphics[
        page=1,
        width=0.8\linewidth,
        trim=3.5cm 0.5cm 3.5cm 0.5cm, 
        clip
    ]{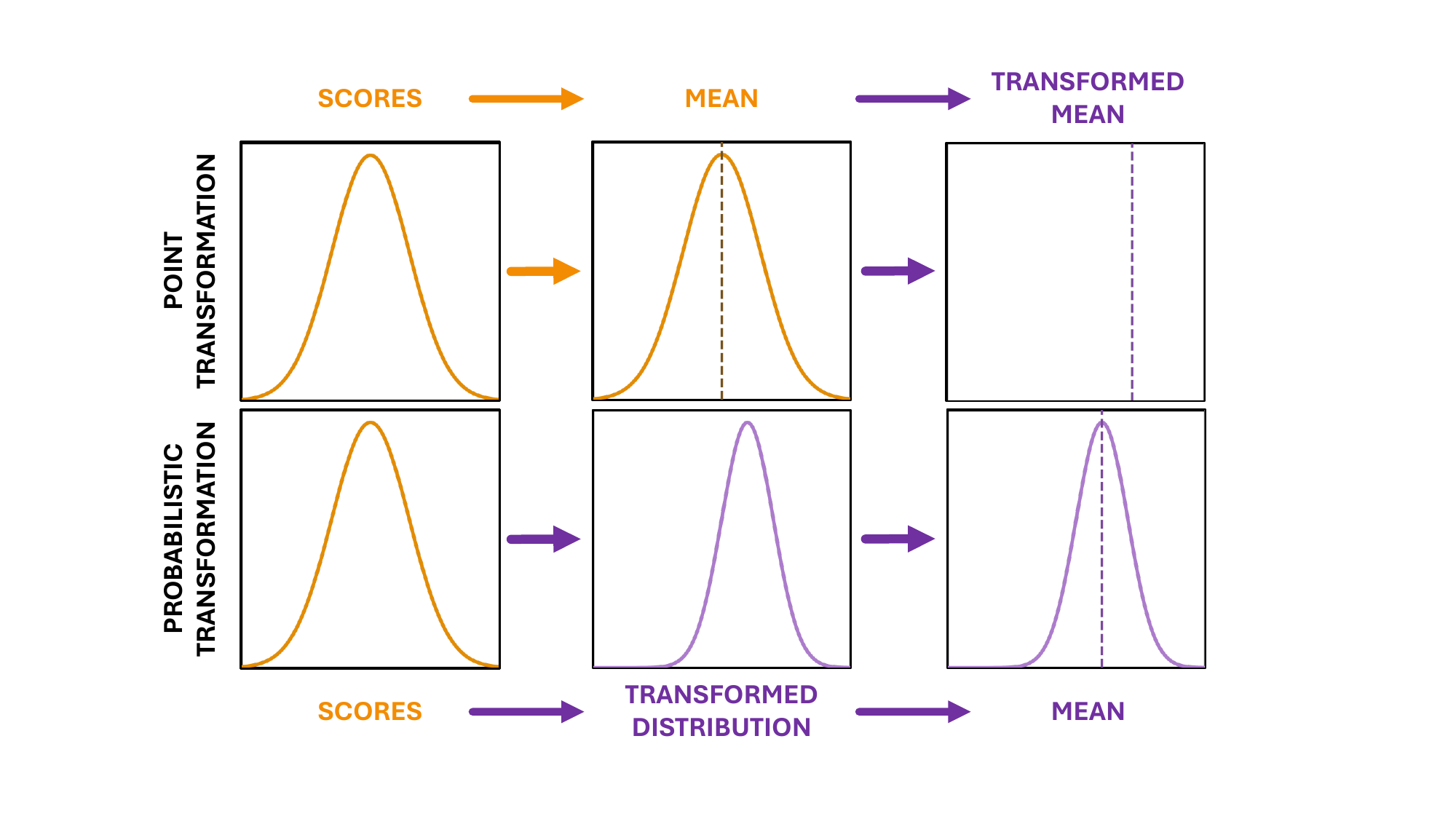}
    \caption{Comparison between point estimation and probabilistic scoring functions.  In the point estimation scoring function, the score distribution is summarized into a single point estimate, which is then transformed.  In contrast, the probabilistic scoring function transforms the entire score distribution using Monte Carlo sampling, and the final value is computed as the mean of the transformed \gls{mpo} scores. Orange represents scores before transformation, while purple represents the transformed scores.}\label{fig:averageVSpoint}
\end{figure}
In REINVENT, most scoring functions return a scalar estimate of the score, and the default implementation applies a non-linear transformation to this point estimate before aggregating the transformed scores into the final \gls{mpo}. However, when a scoring function returns not only a point estimate but a full predictive distribution, as in the case of a ChemProp model trained with \gls{mve}, applying the transformation to the expected value alone is not mathematically consistent.

Because the transformation $\theta_s$ is non-linear and the scoring components are non-deterministic, the current implementation of \gls{mpo} inside REINVENT does not propagate the variance associated with these components:

\begin{equation}
\text{S}_{\text{default}} = f(\mathbb{E}[X]) 
\;\neq\; 
\mathbb{E}[f(X)] = \text{S}_{\text{prob}}
\label{point_average}
\end{equation}

In contrast, a mathematically consistent approach treats the scoring components as random variables and propagates their distributions through the \gls{mpo} calculation (\textit{Fig.}~\ref{fig:averageVSpoint}) \cite{Dodds2024_RLProb}. In this setting, the \gls{mpo} itself becomes a random variable, whose distribution can be approximated using Monte Carlo sampling by evaluating the \gls{mpo} $M$ times. The final \gls{mpo} score is then estimated as the mean over these samples:

\begin{equation}
\mathbb{E}[f(X)] \approx \frac{1}{M} \sum_{m=1}^{M} f(x_m),
\label{AverageDefinition}
\end{equation}

\subsection{Strategies for steering generation toward reliable predictions} \label{MethodsStrategies}

\begin{figure}[!htbp]
    \centering
    \includegraphics[
        page=1,
        width=0.8\linewidth,
        trim=3cm 6cm 3cm 5cm, 
        clip
    ]{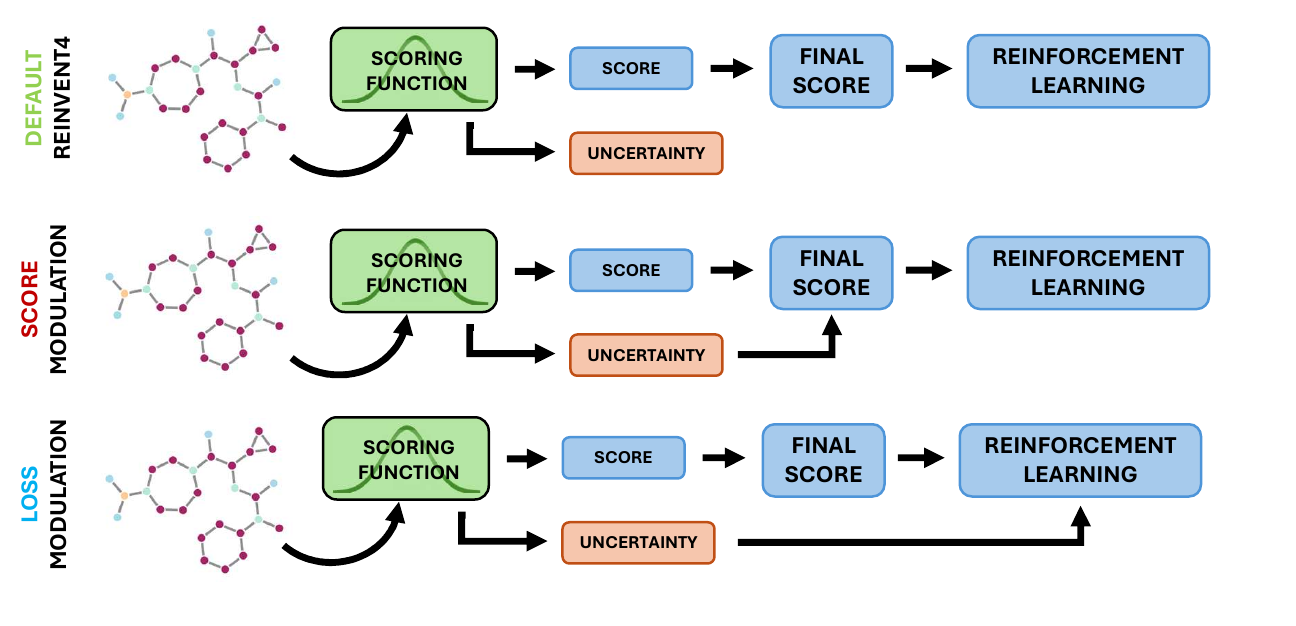}
    \caption{Schematic representation of the different strategies for incorporating uncertainty in \gls{rl}. The top panel shows the default REINVENT4 approach, which ignores uncertainty in predictions. The Score Modulation (SM) strategy includes uncertainty as an additional component in the final scoring function, whereas the Loss Modulation (LM) strategy modulates the contribution of each molecule during the gradient updates.}\label{fig:strategiesEsquema}
\end{figure}

While generative models are capable of exploring vast regions of chemical space, including poorly characterized areas, that are far from the training distribution, this broad exploration is not always desirable. In many practical scenarios, the goal is to generate high-scoring molecules that lie within the reliable prediction region of the scoring function, rather than to explore uncertain regions where predictions may be unreliable. To this end, we investigated two strategies for incorporating predictive uncertainty  into the \gls{rl} framework, steering generation toward low-uncertainty regions: \gls{sm} and \gls{lm}(\textit{Fig.}~\ref{fig:strategiesEsquema}).

In the \gls{sm} approach, uncertainty is treated as an additional optimization objective. Specifically, the predictive uncertainty is incorporated directly into the \gls{mpo} scoring function ($S_{SM}$, Eq.~\eqref{ScoreModulation}). In other words, uncertainty becomes part of the \gls{mpo} objective, encouraging the agent to optimize not only the predicted activity but also the confidence in that prediction.

\begin{equation}
S_{SM}(x_j) = MPO\big(s_1(x_j),\ldots, s_K(x_j), s_{\mathrm{unc}}(x_j)\big)
\label{ScoreModulation}
\end{equation}

In the second approach, uncertainty is not included in the \gls{mpo} scoring function itself ($S_{LM}$, Eq. ~\eqref{LossModulationScore}). Instead, the uncertainty is used to modulate the learning signal during policy optimization. The uncertainty associated with each sample weights its contribution to the policy update  ($L_{LM}$, Eq. ~\eqref{LossModulationLoss}). In other words, molecules contributions are weighted inversely to their uncertainty during gradient updates, controlling how strongly the agent learns from each sample.

\begin{align}
S_{LM}\left(x_j\right) &= MPO\left(s_1\left(x_j\right),\ldots,s_K\left(x_j\right)\right)
\label{LossModulationScore}\\
\mathcal{L}_{LM}(X) &= \frac{1}{N} \sum_{j=1}^{N}\frac{w^{unc}_j}{\frac{1}{N}\sum_{l=1}^{N} w^{unc}_l} \mathcal{L}_j
\label{LossModulationLoss}
\end{align}

When incorporating multiple scoring components with their own independent  uncertainty estimate, we need to aggregate them. In the \gls{sm} strategy,  because uncertainty components are incorporated into the \gls{mpo}, they are aggregated following its predefined formulation (Eq.~\ref{MPO}). Instead, in the \gls{lm} strategy, the uncertainty weights are combined as an arithmetic mean to prevent extreme values from dominating the policy update.

We evaluated each strategy independently: \textit{Noisy Component+\gls{sm}} (\textit{Orange}), \textit{Noisy Component+\gls{lm}} (\textit{Blue}), and the combination of both, \textit{Noisy Component+\gls{sm}\&\gls{lm}} (\textit{Plum}), compared to the \textit{Noisy Component} baseline and, when possible, to a perfect \textit{Oracle}.

\subsection{Experimental setups}\label{Models}
To evaluate the different uncertainty-aware strategies we considered three experimental setups. The first one (i) consists of a controlled Model System with analytically defined uncertainty. Additionally, we assessed performance in two real-data scenarios, (ii) we evaluated a molecular property prediction task using ChemProp models and (iii) we considered a setup in which our uncertainty was derived from a conformal prediction model on top of a \gls{rf} classifier.

\subsubsection{Model system (MS)}\label{MethodsModelSystem}

Our Model System was designed to simulate an ideal \gls{ad} scenario. By definition, within an \gls{ad}, the reliability of predictions is expected to decrease as the distance from the training data increases. Accordingly, we explicitly enforced this behavior by constructing a system in which predictive accuracy degrades when increasing the distance from the training dataset. This was achieved by introducing artificially generated, distance-dependent Gaussian noise to RDKit-computed functions that serve as oracles. This setup allowed us to still retain access to ground-truth values for any sample.

The noise magnitude was defined as a function of the distance between a sampled molecule and the training dataset. To construct the Model System, two different well-known molecular descriptors were used: \textit{logP} and \textit{bertzCT}. Both are continuous and can be computed using RDKit.

The \textit{logP} descriptor corresponds to the octanol-water partition coefficient and measures the hydrophobicity of a molecule. The \textit{bertzCT} descriptor quantifies complexity of a molecule, taking into account complexity of the bond connectivity and the hetero-atom distribution.

 For each sampled molecule $j$, the noisy prediction is given by:

\begin{equation} y_j^{\text{pred}} = y_j^{\text{RDKit}} + \varepsilon_j \label{gaussianNoise} \end{equation}

where

\begin{equation} \varepsilon_j \sim \mathcal{N}\!\left(0, \sigma_j^2\right), \quad \sigma_j = a + b \cdot d_j \label{GaussianDefinition} \end{equation}

Here, the standard deviation $\sigma_j$ is defined as a function of the distance $d_j$ to the training data (Eq.~\eqref{distanceDefinition}). The parameter $a$ represents a baseline noise floor that is present even for molecules within the training domain, while $b$ controls how rapidly noise increases with distance. In our experiments, we set $a = 0.1$ and $b = 0.9$.

This construction ensured that predictions for molecules close to the training data are approximately accurate, whereas predictions for distant molecules are increasingly corrupted by noise, thereby simulating a controlled \gls{ad} scenario with known ground-truth values.

Several distance measures were evaluated, including Tanimoto similarity. However, during extensive sampling with REINVENT4, this metric exhibited a limited range variation. Therefore, instead, Euclidean distance was used within a \gls{pca}-transformed space. In this context, the specific choice of distance metric was not critical, as the objective was not to assess the informativeness of the uncertainty measure, but rather to obtain a distance that meaningfully varies throughout the chemical space and is consistent with the concept of \gls{ad}. 

The \gls{pca} model was constructed for each scoring function using Morgan fingerprints (radius $r=3$, $2048$ bits) computed for the training dataset of the respective noisy scoring components: $\gls{egfr}_{\text{training}}$ was used for \textit{logP} and $\gls{drd2}_{\text{training}}$ for \textit{bertzCT}. These high-dimensional binary vectors were then used to fit a \gls{pca} model with $n=10$ components.

This \gls{pca}-transformed space defined the applicability domain: molecules that are close to the training projections in this space are considered to lie within the domain, while those that are distant are considered out-of-domain. All subsequent distance calculations were performed in this \gls{pca} space.

During \gls{rl}, each sampled \gls{smi} was converted into the same fingerprint representation used for the training set, namely a fingerprint with radius $r=3$ and $2048$ bits, and then projected using the fitted \gls{pca} transformation. For each generated sample, we computed the mean Euclidean distance, denoted by $\mu_j$, to its $k=5$ nearest neighbors $\mathcal{N}_k(j)$ in the PCA-transformed training set:

\begin{equation}
\mu_j = \frac{1}{k} \sum_{l \in \mathcal{N}_k(j)}
\left\| \mathbf{z}_j - \mathbf{z}_l \right\|_2
\label{neighboursDefinition}
\end{equation}

Additionally, a correction term based on the standard deviation of the $k$-nearest neighbor distance, $\sigma^{knn}_j$, was included in the definition to account for local variability. During the experiments we used 5-nearest neighbors.

The final distance measure is defined as:

\begin{equation}
d_j =\theta_{dist}( \mu_j + \lambda \sigma^{knn}_j)
\label{distanceDefinition}
\end{equation}

We transformed the raw distances into a score making use of the following $\theta_{dist}$ sigmoid transformation centered at an inflection point $\beta$:

\begin{equation}
    \theta_{dist}(d) = \left(1 + \exp\bigl(-\alpha\,(d - \beta)\bigr)\right)^{-1}
    \label{eq:sigmoidDistances}
\end{equation}

We evaluated two configurations that differ slightly in these parameters (see \textit{\gls{si} Section~6} for full parameter derivations). The first, \textit{mid-initial-distances}, simulated a broad applicability domain in which majority of the randomly sampled molecules by the Reinvent prior fell in an intermediate uncertainty region, resulting in a mean initial distance of $\bar{d}_0 \approx 0.5$. The second, \textit{high-initial-distances} setting, shifted the distribution of randomly sampled distances to $\bar{d}_0 \approx 0.8$, imposing a tighter boundary where most randomly sampled molecules already lie in high-uncertainty regions, penalizing out-of-distribution molecules earlier.

In the Model System, the final uncertainty-based weights ($w^{unc}_j$) were defined as the linear transformation of the distance, given by:

\begin{equation}
w^{unc}_j=1 - d_j
\label{simDefinition}
\end{equation}

These weights were used in both strategies: as an additional scoring component in the \gls{sm} strategy and as the weights to rescale the policy updates contributions in the \gls{lm} strategy.

\subsubsection{ChemProp Models} \label{MethodsChemProp}

To train the ChemProp models, we made use of the previously introduced datasets (\textit{Section~\ref{MethodsDatasets}}), $\gls{egfr}_{\text{small}}$ and $\gls{egfr}_{\text{full}}$.

Two different ChemProp models were constructed, a model trained on the $\gls{egfr}_{\text{full}}$ dataset, that served as the ground-truth reference and was referred to as the \textit{ChemProp Oracle}. Additionally, a predictive model was trained on $\gls{egfr}_{\text{small}}$, which is a substantially smaller dataset. This model was used as the scoring function during optimization and is referred to as \textit{ChemProp Predictor}.

Both models were implemented using ChemProp and were trained using \gls{mve}, enabling the model to predict both the score and an associated predictive uncertainty. Further details on the ChemProp model training is provided in the \textit{\gls{si} Section~3}. 

To improve the reliability of the uncertainty estimates, a calibration procedure was applied using the test set. The calibration was performed by fitting an \gls{mve}-based weighting calibrator.

These ChemProp models yield two outputs: the activity prediction, which was used to guide the optimization objective, and the associated \gls{mve}-based uncertainty estimate, which was used as a measure of predictive uncertainty. This uncertainty was transformed into an uncertainty-based weight ( $w^{unc}_j$ ) analogous to the similarity-based weight in Eq.~\eqref{simDefinition}, such that lower uncertainty was assigned higher weight according to $ w^{unc}_j = 1 / \mathrm{uncertainty}$.

\subsubsection{Conformal prediction models} \label{MethodsConformalPredictors}
We constructed a binary classification model to predict molecular activity using a \gls{rf} classifier, with uncertainty quantification provided by a Mondrian \gls{icp} \cite{Norinder2014_MondrianICP}. Molecular structures were encoded as \gls{ecfp}  using RDKit with a radius $r=3$ and $nBits=2048$. The dataset used to construct the classifier was the training dataset $\gls{egfr}_{\text{full}}$. The \gls{rf} classifier (500 trees, scikit-learn implementation) was trained on the training set, to distinguish between actives and inactive compounds with a threshold activity of  $pIC_{50} \geq 6$. 

The \gls{cp} was implemented using the nonconformist package \cite{linusson_nonconformist}. We made use of \gls{icp}, calibrating the nonconformity scores on the test set. The nonconformity score was defined using the margin error function:
\begin{equation}
    \alpha_i = \frac{1}{2} - \frac{\hat{P}(y_i \mid x_i) -
    \max_{y' \neq y_i} \hat{P}(y \mid x_i)}{2}
    \label{eq:nonconformity}
\end{equation}

where $\hat{p}(y \mid x)$ is the \gls{rf} class probability estimate.
For each molecule, the p-value for class $y$ is computed as the fraction of calibration
nonconformity scores that are at least as large as the score computed under the hypothesis that
the test label is $y$:

\begin{equation}
    P(y) = \frac{\left|\left\{i \in \mathcal{D}^{y}_{\text{cal}} :
    \alpha_i \geq \alpha(x_{\text{test}}, y)\right\}\right| + 1}
    {|\mathcal{D}^{y}_{\text{cal}}| + 1}
    \label{eq:pvalue}
\end{equation}

where $\mathcal{D}^{y}_{\text{cal}}$ is the calibration subset with true label $y$.

For each compound, the \gls{icp} outputs a prediction set (active, inactive, or both) at a
specified confidence level (95\% in our experiments).

We used the conformal p-value associated with the active class ($p_{\text{active}}$) as the continuous measure of predicted activity, with higher values indicating a higher likelihood of activity.
Uncertainty was quantified using the second-largest class p-value (serving as a proxy of efficiency), which in the binary setting is $\min(p_{\text{active}},\, p_{\text{inactive}})$: a high value indicates an ambiguous prediction set that includes both classes.
This uncertainty measure was transformed into a weight as:

\begin{equation}
    w^{\text{unc}}_j = 1 - \min\!\left(p_{\text{active},j},\, p_{\text{inactive},j}\right)
    \label{eq:uncertainty_weight}
\end{equation}

\subsection{Optimization objectives}\label{MethodsOptimizationObjectives}
In addition to the primary scoring function under evaluation (\textit{LogP}, \textit{BertzCT}, ChemProp activity prediction, or conformal prediction models), we included \gls{mw} and \gls{tpsa} as additional optimization objectives.

All scoring components were transformed using sigmoid or double-sigmoid functions to favor values within predefined optimal ranges. The specific parameterization of each transformation is provided in \textit{\gls{si} Section~2}.

To promote chemical diversity during optimization, we applied the Identical Murcko Scaffold diversity filter already implemented in REINVENT. This filter penalizes the repeated generation of compounds sharing the same Bemis–Murcko scaffold once a predefined bucket size is exceeded, encouraging the exploration of novel scaffolds, and was included to mimic real use-cases.

\subsection{Evaluation metrics}\label{MethodsEvaluationMetrics}

Model performance was primarily assessed by the number of unique hit-scaffolds. Hits-scaffolds are defined as molecules with unique Bemis-Murcko scaffolds that achieve a transformed score above $0.65$ for every individual scoring component (i.e., \gls{mw}, \gls{tpsa}, and all activity predictions used in the given experiment).

When an oracle was available, we further distinguished between total predicted accumulated hit-scaffolds and true hits, where true hits were defined as compounds confirmed to be active by the oracle. For the \gls{cp} setting, where no oracle was available, true hits were instead defined as compounds meeting the hit criteria and assigned exclusively to the active class by the \gls{cp} (that is, prediction set size 1).

Additionally, results were analyzed in the \gls{pca} space to visualize the distribution of generated compounds relative to the training data when relevant.


\section{Results}\label{Results}
\subsection{Results for Model System}

\subsubsection{One noisy component}

We first consider the Model System with a single noisy component where the RDKit-computed \textit{logP} serves as the reference score. Gaussian noise drawn from \textit{Eq.~\eqref{GaussianDefinition}} is added to the oracle, as described in \textit{Section~\ref{MethodsModelSystem}}. In this setup the magnitude of the noise aligns with the distance metric as previously defined in \textit{Eq.~\eqref{distanceDefinition}}. Consequently, the distance against the training dataset can be interpreted as an uncertainty measure, simulating an \gls{ad}-type uncertainty.

\begin{figure*}[!htbp]
    \centering
    \includegraphics[page=1,width=0.8\linewidth]{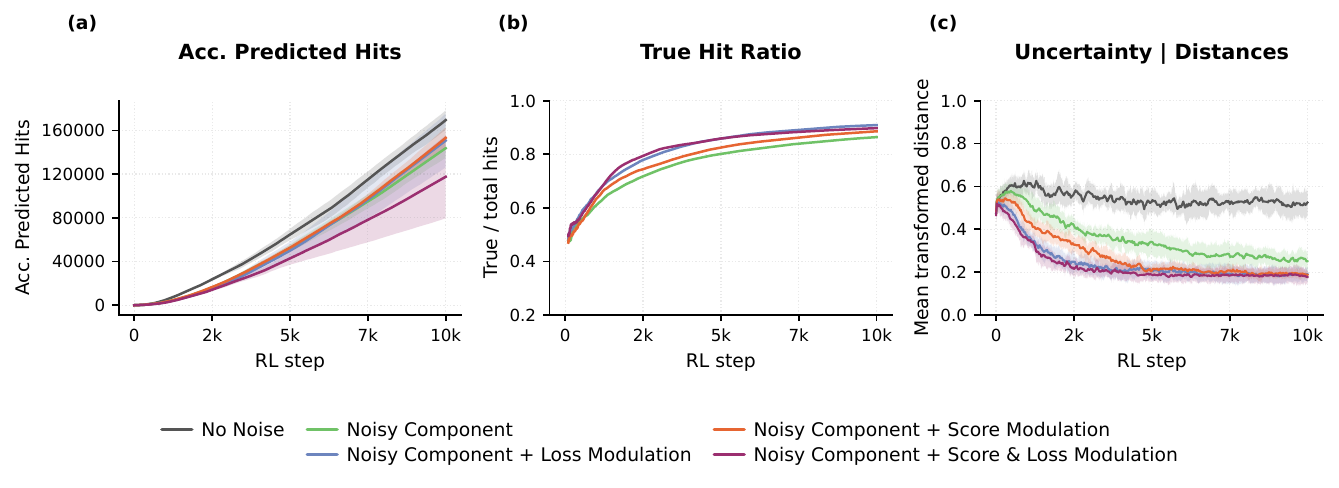}
    \caption{Results for Model System with one noisy scoring component using RDKit \textit{logP} as the scoring predictor. We report (a) the total number of accumulated hit-scaffolds, (b) the True hit ratio and (c) the mean transformed distance, which in Model System corresponds to the uncertainty measure. For all reported scores, except uncertainty, values closer to 1 indicate better performance. Results are averaged over five independent runs, with the standard deviation shown as the shaded area. For an extended figure see \textit{\gls{si} Section~4}.}
    \label{fig:resultsHits_Mid}
\end{figure*}

In the absence of noise (\textit{No noise}, \textit{Black}), \textit{logP} acts as a perfect oracle. Under this condition, no guidance relative to the training dataset is imposed, and the model exploration is not biased towards chemically similar regions. The model explores the molecular space broadly, covering both regions close and far away from the training data,  while maintaining a mean transformed  distance close to 0.5 (\textit{Fig.}~\ref{fig:resultsHits_Mid}~(c)), consistent with the initialization procedure described in \textit{Section~\ref{MethodsModelSystem}}. In other words, in this situation the prior is able to explore and sample molecules from a broad region of the chemical space, including regions that would be classified as high uncertainty (the ones far away from the dataset) and low uncertainty (the ones close to the training dataset).

In contrast, when introducing distance-dependent noise ($\mathcal{N}(a=0.1, b=0.9)$, \textit{Noisy Component, Green}), we observe a different behavior. Even without explicitly incorporating uncertainty guidance, the model exhibits an implicit corrective mechanism, in which the presence of noise shifts the exploration towards regions of low uncertainty (\textit{Fig.}~\ref{fig:resultsHits_Mid}~(c)). This suggests a form of self-regularization, where highly noisy regions are avoided due to instability in the scoring signal, observing an empirical tendency to move toward lower-uncertainty regions even without explicit uncertainty guidance. Although this behavior is  potentially beneficial in real optimization settings, it complicates our analysis, making it harder to differentiate between the intrinsic effect and the impact of our explicit uncertainty guidance strategies.

We observed that the self-regularization effect depends strongly on the initial distribution of sampled molecules. When the initial samples lie far from the training data (e.g., with an initial average distance of approx. $\bar{d} \approx 0.8$), the self-regularization effect is significantly weakened. In such cases, majority of the sampled molecules reside in high-uncertainty regions, such that the model’s guiding signal toward more stable regions is weaker, requiring substantially more epochs before a clear shift becomes observable (see \textit{\gls{si} Section~5\&6}).


Nevertheless, both uncertainty-guided reinforcement learning strategies, \gls{sm} and \gls{lm}, induce a faster shift toward lower-uncertainty regions, reducing the number of epochs required, compared to the model’s intrinsic self-correction mechanism.

While differences in the total number of accumulated hit-scaffolds between different methods are relatively small,  when making use of the uncertainty guidance strategies, we observe a decrease in accumulated false hits and therefore, the ratio of true to total hits shows a small improvement, particularly compared to the \textit{Noisy component} by itself (\textit{Fig.}~\ref{fig:resultsHits_Mid}, for an extended figure see \textit{\gls{si} Section~4}).

When using the \gls{sm} strategy (\textit{Orange}), the model transitions into low-uncertainty regions more rapidly than the \textit{Noisy Component}  baseline \textit{(Green)}, reaching a mean distance of 0.2 by 5K steps, a minimum that the \textit{Noisy Component} alone does not achieve (see also distance distributions in \textit{\gls{si} Section~6}). Since in the Model System uncertainty aligns perfectly with prediction error, this reduction in distance directly translates into more accurate scores. This is reflected in a small increase in mean predicted \textit{logP} values and nearly halving the number of false hits (\textit{\gls{si} Section~4 (b)}), consequently slightly improving the True/Total hit ratio.

With the \gls{lm} strategy (\textit{Blue}), we observe a further improvement over \gls{sm}. This strategy yields a better performance not only in terms of uncertainty, but also obtaining a significant increase in the ratio True/Total unique hit-scaffolds. 

Interestingly the combined approach (\gls{sm}\&\gls{lm}, \textit{Plum}), although the shift into highly similar regions of the space is even faster,  the number of total accumulated hit-scaffolds is significantly lower in multiple experiments (\textit{Fig.}~\ref{fig:resultsHits_Mid}, \textit{\gls{si} Section~4\&5}).

Analysis of the distributions of the uncertainties throughout the \gls{rl} runs supports the idea of this self-correction, as evidenced by a shift toward lower-uncertainty regions, not seen in the \textit{No noise} case, which uncertainties spreads throughout the whole landscape. Although, we still observe that this shift into low uncertainty regions is more pronounced when using uncertainty-guided methods (\textit{\gls{si} Section~6}).

\subsubsection{Multiple noisy components} 

When using two noisy components in the Model System, we introduce two different continuous RDKit descriptors that are weakly correlated with each other (\textit{LogP} and \textit{BertzCT}). The rationale is to construct two different objectives that drive the exploration toward different regions of chemical space. 
\begin{figure*}[!htbp]
    \centering
    \includegraphics[page=1,width=0.8\linewidth]{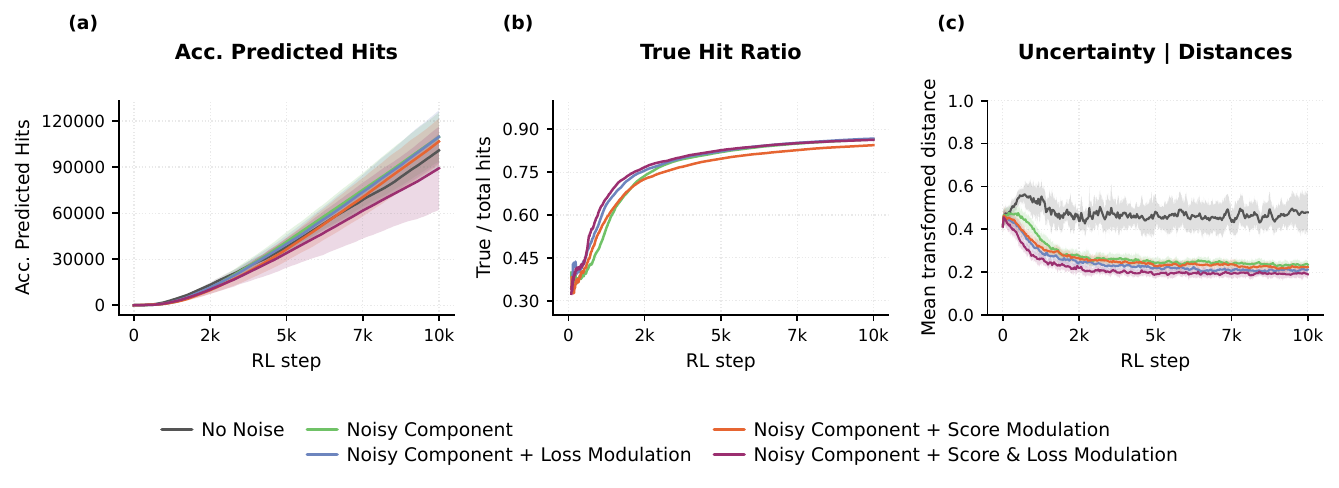}
    \caption{Results for Model System with two noisy scoring component using RDKit \textit{logP} and RDKit \textit{BertzCT} as the  scoring predictors. We report (a) the total number of accumulated hit-scaffolds, (b) the ratio of true to total accumulated hit-scaffolds. Additionally, we report (c) the geometric mean of the transformed distances, which in Model System corresponds to the uncertainty measure. For all reported scores, except uncertainty, values closer to 1 indicate better performance. Results are averaged over five independent runs, with the standard deviation shown as the shaded area. For an extended figure see \textit{\gls{si} Section~7}.}
    \label{fig:resultsHits_Mid_Multi}
\end{figure*}
The definition of noise is extended in an analogous manner to the single-component setting. Each RDKit descriptor is perturbed independently using Gaussian noise, with a variance that scales with the corresponding distance metric to its respective training dataset, as defined previously in Eq.~\eqref{GaussianDefinition} and Eq.~\eqref{distanceDefinition}.

Although we made use of two different independent datasets targeting \gls{egfr} ($\gls{egfr}_{\text{training}}$ for \textit{logP} scoring function) and \gls{drd2} ($\gls{drd2}_{\text{training}}$ for \textit{bertzCT} scoring function), in practice, both datasets consist of active compounds against human receptors and therefore belong to a shared drug-like chemical space (i.e., Lipinski-compliant region). This overlap allows the identification of regions where uncertainty can be jointly reduced across both components.

As in the single-component case, we do not observe substantial differences in total number of accumulated hit-scaffolds across the different methods. In this case we do not see significant differences in terms of false hits. As a consequence the ratio True/Total hit-scaffolds it is really similar (\textit{\gls{si} Section~7}). 

Although the oracle RDKit descriptor (\textit{“No Noise”}) baseline remains centered with a distance close to 0.5, the noisy scoring function, without any explicit uncertainty correction, drives the exploration toward regions of lower uncertainty for both datasets. This correction is even more marked than in the unique noisy component. As a consequence, differences in the total number of hit-scaffolds and false hits become negligible.

In additional experiments, we modified $\beta$ of the transformation in Eq.~\eqref{distanceDefinition} to produce a higher initial mean distance, thereby increasing the mean uncertainty of randomly sampled molecules using REINVENT. Under these conditions, the self-correction effect is still observed; however, substantially more optimization steps are required before the exploration reaches low-uncertainty regions. In this setting, clearer differences emerge between our proposed methods and the noisy-component-only. Specifically, we observe a reduction in the number of false hits together with an increase in the true-to-total hit ratio. Once again, the combined \gls{sm}\&\gls{lm} approach shifts exploration toward lower-uncertainty regions, but at the cost of a substantially reduced number of hit-scaffolds, ultimately leading to a detrimental effect on overall performance (\textit{\gls{si} Section~8}).


\subsection{Results for real-data noisy component - ChemProp models}

We evaluate the different strategies using the \textit{ChemProp Predictor} model trained on the $\gls{egfr}_{\text{small}}$ dataset alongside the \textit{ChemProp Oracle} trained on the $\gls{egfr}_{\text{full}}$ dataset. Both models are trained using the same protocol and exhibit similar behavior; the only distinction lies in the size of the training data, as described in \textit{Section~\ref{MethodsChemProp}}.

\begin{figure*}[!htbp]
    \centering
    \includegraphics[page=1,width=0.8\linewidth]{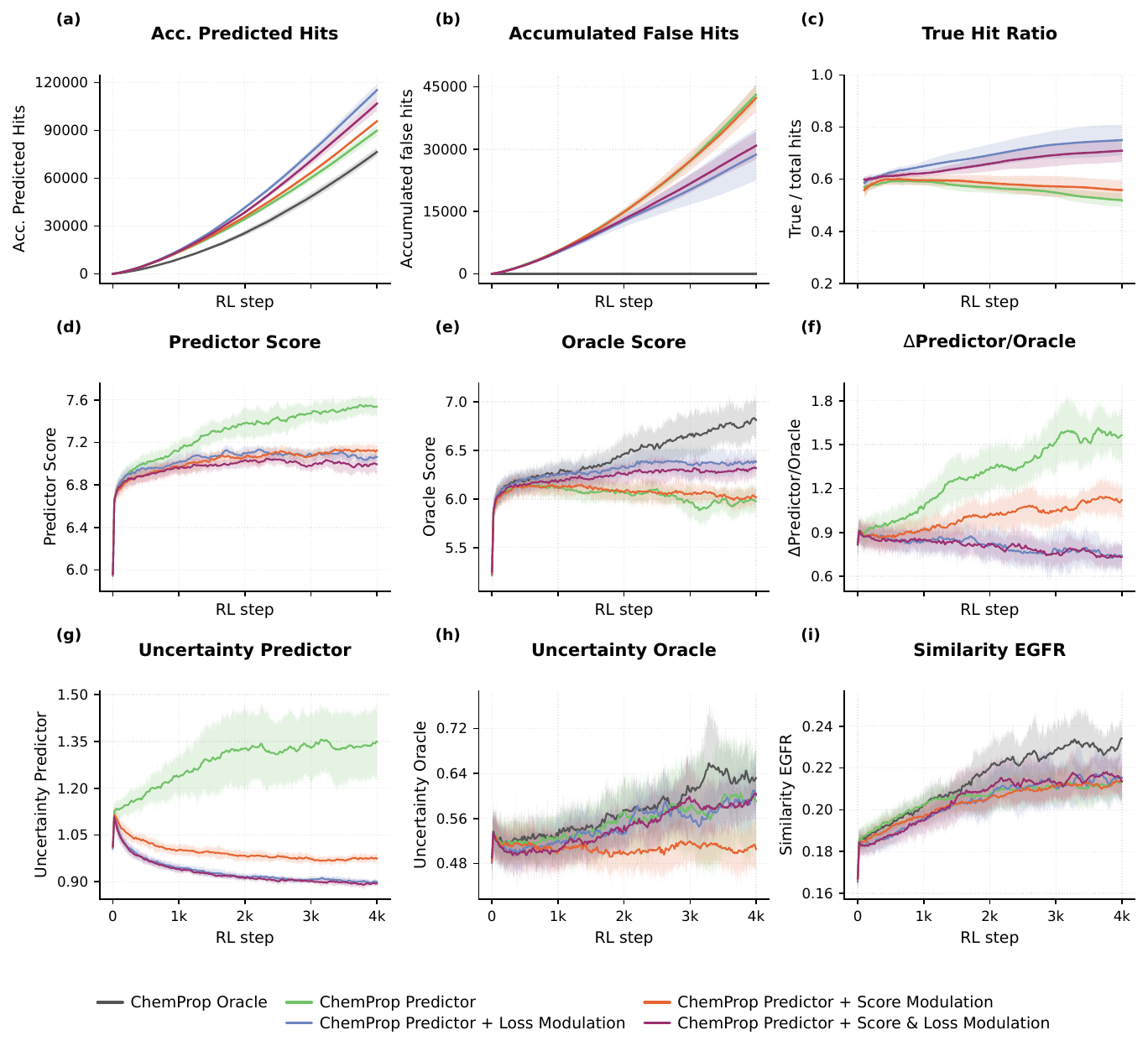}
    \caption{Results obtained using ChemProp Predictor model as the activity scoring predictor. We report (a) the total number of accumulated hit-scaffolds, (b) the number of false hits among them, and (c) the ratio of true to total accumulated hit-scaffolds. Additionally, we report several metrics throughout the \gls{rl} run, including (d) the \textit{ChemProp Predictor} score use to guide the optimization, (e) the \textit{ChemProp oracle} score representing the the ground-truth score, (f) the difference between \textit{Predictor} and Oracle, reflecting prediction error, (g) the \textit{Predictor} model uncertainty, (h) the \textit{ChemProp Oracle} uncertainty and (i) the mean Tanimoto similarity against $\gls{egfr}_{\text{small}}$. For the \textit{ChemProp Predictor} and \textit{Oracle} scores, values closer to 1 indicate better performance, whereas for uncertainty, values closer to 0 indicate higher reliability. Results are averaged over five independent runs, and the shaded area denotes the standard deviation between runs.}
    \label{fig:resultsChemProp}
\end{figure*}

When using these ChemProp models with \gls{mve} prediction, we employed the probabilistic scoring functions introduced in Section~\ref{methods:ProbScoringFucntions}, as these are the mathematically correct approach for probabilistic scoring predictors. A comparison with the corresponding point-estimate scoring functions is provided in \textit{\gls{si} Section~9}.

When optimization is driven solely by the \textit{ChemProp Predictor} without uncertainty guidance (\textit{Green}), the predicted activity score increases strongly and reaches the highest values among all methods. However, this is accompanied by a substantial increase in predictive \gls{mve} uncertainty. Comparison with the \textit{ChemProp Oracle} shows that the \textit{Predictor} consistently overestimates activity, with an average error of approximately 0.9 IC$_{50}$ units, and that this discrepancy increases over the course of the \gls{rl}.
Notably, the increase in predicted activity is not reflected in the \textit{ChemProp Oracle} score, indicating that optimization is being driven toward regions of chemical space that appear favorable to the \textit{Predictor} but in reality are poorly characterized and highly uncertain. The corresponding \textit{Oracle} uncertainty remains low, indicating that these lower \textit{Oracle} scores are not caused by increased uncertainty in the \textit{Oracle} model.

In contrast, introducing uncertainty-guided strategies mitigates this effect, leading to a noticeable reduction in uncertainty. Although they yield more moderate \textit{Predictor} scores, they maintain a smaller and more stable gap between \textit{Predictor} and \textit{ChemProp Oracle} scores, indicating more reliable optimization. Importantly, these \textit{Oracle} scores are likewise associated with low \textit{Oracle} uncertainty, suggesting that uncertainty guidance helps retain optimization within regions that are considered reliable by both models.

Notably, the \textit{Oracle} yields the lowest total number of accumulated unique scaffold predicted hits. This is not unexpected, as our \textit{Oracle} is not a true experimental oracle but rather a stronger surrogate model whose predictions we treat as an upper bound. The larger number of predicted hits by the \textit{ChemProp Predictor} is therefore primarily a consequence of the systematic overestimation relative to the \textit{Oracle}, rather than an evidence of improved scaffold discovery. Given the average overestimation of approximately 0.9 IC$_{50}$ units, this difference in predicted hit counts is expected.

Among the evaluated strategies, \gls{lm} achieves the highest number of predicted hits, followed by \gls{sm}\&\gls{lm}, whereas the no-guidance and \gls{sm} settings remain similar and produce fewer hits. A similar pattern is observed for false hits: despite generating more hits overall, \gls{lm} also yields the fewest false hits, followed by \gls{sm}\&\gls{lm}, while the no-guidance and \gls{sm} settings again perform similarly. Consequently, \gls{lm} produces the largest number of true hits overall, similar to the \textit{Oracle}, yielding the strongest improvement in the true-to-total hit ratio among the guided strategies.

\subsection{Results for conformal prediction as uncertainty measure}
We further evaluate the proposed strategies using a \gls{cp} built on top of a \gls{rf} classifier. In this setting, no external ground-truth oracle is available.

\begin{figure*}[!htbp]
    \centering
    \includegraphics[page=1,width=0.8\linewidth]{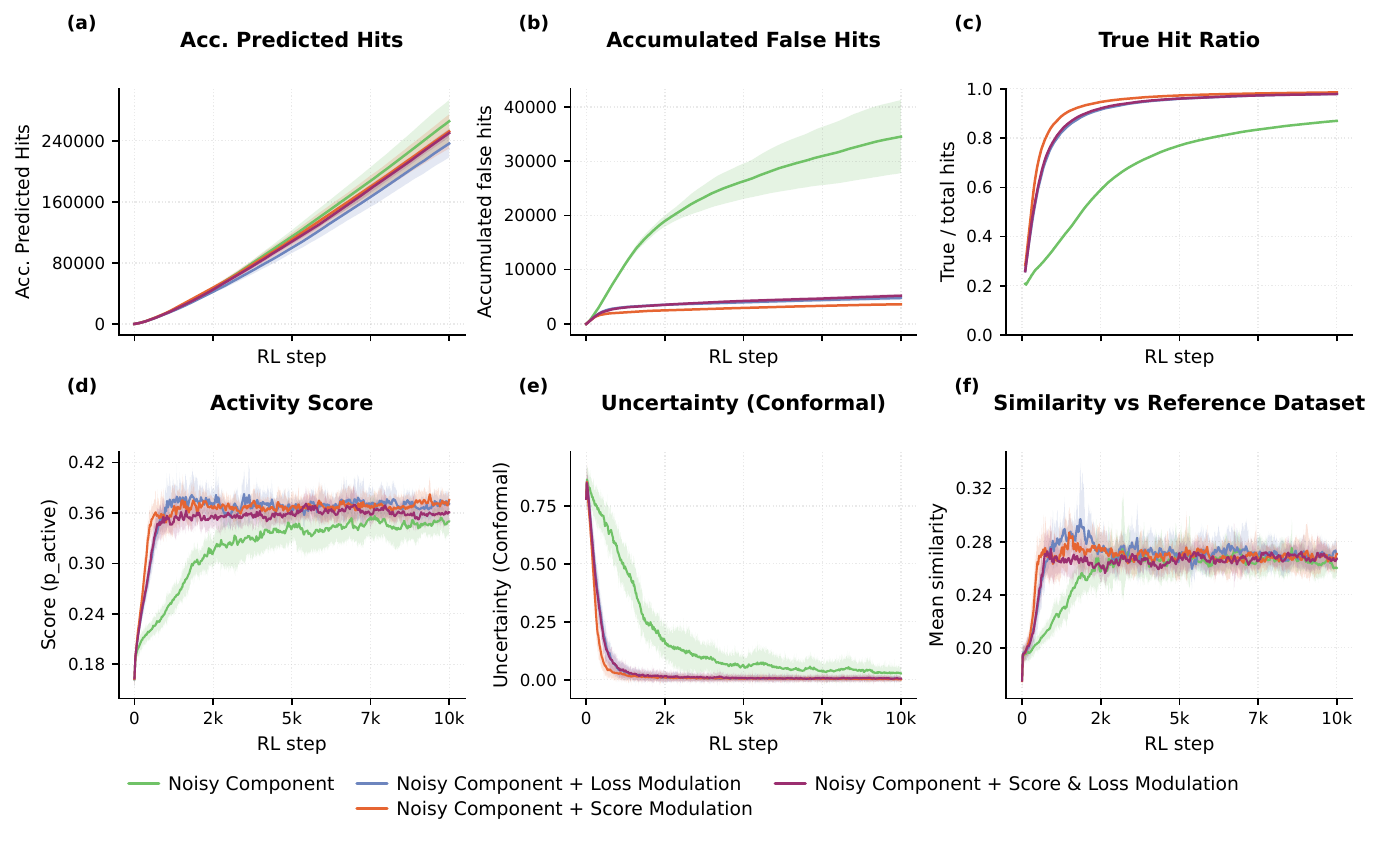}
    \caption{Results obtained using a \gls{cp} built on top of a \gls{rf} classifier as the activity scoring predictor. We report (a) the total number of accumulated hit-scaffolds, (b) the number of false hits, where false hits were defined as hits that are ultimately classified as uncertain by the \gls{cp}, and (c) the ratio of true to total accumulated hit-scaffolds. Additionally, we report several metrics throughout the \gls{rl} run, including (d) the predictive activity score used to guide the optimization process, (e) the uncertainty associated to those predictions, and (f) the mean Tanimoto similarity against $\gls{egfr}_{\text{training}}$. Activity scores closer to 1 indicate better performance, whereas uncertainty values closer to 0 indicate higher reliability. Results are averaged over five independent runs, with the standard deviation shown as the shaded area.}
    \label{fig:resultsConformal}
\end{figure*}

When using only the noisy predictive component (i.e., without uncertainty guidance), we observe that the $p_{\text{active}}$ score increases steadily throughout the \gls{rl} process, while the associated uncertainty decreases. This behavior is accompanied by an increase in the mean Tanimoto similarity with respect to the training dataset, indicating that the model progressively focuses on regions of chemical space that are closer to the training distribution.

When incorporating uncertainty-guided strategies, we observe a similar but more pronounced trend: the $p_{\text{active}}$ score increases to slightly higher values, while the uncertainty decreases more rapidly, reaching the significance level. This behavior is expected, as the conformal predictor is designed such that uncertainty cannot fall below the predefined significance level. Tanimoto similarity also increases faster. 

Although the total number of hit-scaffolds is comparable across all strategies, the uncertainty-guided approaches produce a substantially higher number of compounds assigned a unique class label by the conformal predictor, which we refer to as \textit{true hits}. In contrast, when not using the uncertainty guidance strategies, just using the activity score by itself, generates a much larger proportion of ambiguous predictions, where compounds are simultaneously assigned to both classes by the \gls{cp} (\textit{Fig.}~\ref{fig:resultsConformal}).

\section{Discussion}\label{Discussion}
We have shown how uncertainty-guided strategies can help stabilize exploration of chemical space. By leveraging uncertainty to guide \gls{rl}, the model focuses the exploration on regions of low uncertainty, around which the scoring functions are more reliable, without a detrimental effect on score. This effect is evident not only in the Model System setting, where the uncertainty measure is designed to be perfect and closely aligned with prediction error, but also in more realistic scenarios such as the ChemProp models and the \gls{cp} model.

Interestingly, in the Model System setting, we observe a form of self-regularization, where the model naturally shifts towards less uncertain regions of the space. This behavior is intuitive: regions associated with high uncertainty are unstable, leading to noisy learning signals. In contrast, low-uncertainty regions provide more consistent predictions and therefore the learning signal is more stable. This interpretation is further supported by the observed reduction of the self-regularization effect as the noise variance decreases. Since our Model System contains controlled Gaussian noise, lowering the variance reduces the instability of these regions.

Additionally, we observe that this self-regularization effect is highly dependent on the initial landscape distribution of uncertainties. In a situation in which the  distance transformation becomes broader, and we have substantially less frequent low uncertainty compounds, the self-regularization becomes less effective. The optimization landscape becomes more difficult, and just the self-regularization is not enough to drive this shift, although when including our uncertainty-aware strategies we are able to reach and explore these low uncertainty regions. 

In real-world scenarios, however, this self-regularization effect becomes substantially less effective. Although it remains observable in the ChemProp models ((h) in Fig.~\ref{fig:resultsChemProp}), it is insufficient to prevent the model from converging toward regions associated with high uncertainty, ultimately leading to a larger overall prediction error. 

When sampling molecules from high scoring and at the same time high uncertainty regions, when comparing the \textit{Predictor} model to the \textit{Oracle}  score, we observe little correlation. In other words, although the \textit{Predictor} score increases, the \textit{Oracle} performance remains low. This highlights the importance of explicitly incorporating uncertainty guidance during optimization. In reality, these regions are not truly high-scoring but are instead poorly characterized, leading to wrong predictions. 

In contrast, when using our \gls{lm} strategy in the same real-task setting with the \textit{ChemProp} models, the \textit{Predictor} scores exhibit a substantially stronger correlation with the \textit{Oracle} score. As a result, increases in the \textit{ChemProp Predictor} score are also reflected as an increase in the ground-truth values represented by the \textit{ChemProp Oracle}. This improvement is achieved by steering the optimization toward regions of lower predictive uncertainty. Consequently, this approach not only increases the total number of hit-scaffolds, but also improves the ratio of true to total hit-scaffolds by nearly 0.3, resulting in almost twice as many true hits overall.

For the \gls{cp} model, the absence of ground truth prevents a direct evaluation of prediction accuracy. Nevertheless, we have used as proxy of \textit{True hits} the subset of molecules that are both classified as hits and assigned a unique class label by the conformal predictor. Although we observe that all methods produce a similar number of accumulated hits, with no significant differences between strategies, when uncertainty guidance is not used, we observe a substantial increase in the number of molecules assigned to both classes by the conformal predictor. 

In contrast, both uncertainty-guided strategies consistently reduce the associated uncertainty. Importantly, this reduction is accompanied by an increase in the True/Total hit ratio, indicating that a larger fraction of the generated hits correspond to confident predictions. Additionally, we observe that decreasing uncertainty tends to produce a slight increase in similarity against the training dataset, suggesting that the optimization progressively shifts toward regions of chemical space that are better represented in the training distribution.

Across all considered scenarios, the incorporation of uncertainty guidance consistently improved performance, supporting its inclusion in the \gls{rl} optimization process whenever a meaningful estimate of predictive uncertainty is available. While our comparison does not include all possible settings, the results suggest that the \gls{lm} strategy provides the most favorable overall performance, achieving lower uncertainty while also yielding a larger number of hits.

We hypothesize that the superior performance of the \gls{lm} strategy arises from the way in which uncertainty is incorporated into the optimization process. 

In the case of \gls{sm}, uncertainty is directly integrated into the \gls{mpo} score itself. As a consequence, the learning signal becomes a combination of the original real score and the uncertainty estimate. However, uncertainty is not a score itself, but rather a measure of confidence in the prediction. Therefore, incorporating it directly into the optimization objective can distort the reward landscape and potentially drive the model toward directions that do not align with the underlying scoring objective.

In contrast, the \gls{lm} strategy preserves the original scoring function intact and instead uses uncertainty to modulate the contribution of individual samples to the policy updates. Molecules associated with lower uncertainty contribute more strongly, whereas uncertain molecules have a reduced influence. Importantly, this approach does not alter the meaning of the reward itself, but instead controls how strongly each sample affects learning. As a result, \gls{lm} provides a more natural and stable mechanism for incorporating uncertainty, allowing the model to prioritize reliable learning signals without explicitly biasing the optimization objective away from the intended scoring function.

When considering the combined approach, \gls{sm}\&\gls{lm}, in the real-data scenarios we observe a performance comparable to that of \gls{lm} alone. In the Model System, however, we observe that the shift toward low-uncertainty regions of the chemical space occurs earlier in the optimization process compared to other strategies. Interestingly, rather than yielding a greater number of unique scaffold hits, this accelerated shift leads to a reduction in the number of hits. A closer inspection reveals an increase in the number of scaffolds encountered more than 25 times, a metric associated with our diversity filter, accompanied by a slight decrease in the fraction of valid \gls{smi}. We hypothesize that in the Model System, where the uncertainty measure is perfectly calibrated and closely aligned with prediction error, the combined strategy applies a redundant and compounding pressure toward low-uncertainty regions. This over-correction accelerates convergence at the expense of diversity. In real-data scenarios, where uncertainty estimates are inherently noisier, this effect is attenuated, although a more detailed analysis of hit diversity remains an important direction for future work.

This work demonstrates that incorporating predictive uncertainty into the \gls{rl}-driven molecular generation process yields consistent and meaningful improvements across diverse experimental settings. By guiding the optimization away from poorly characterized regions of chemical space, uncertainty-aware strategies reduce the risk of exploiting spurious predictions while increasing the reliability of generated hits. Among the strategies evaluated, the \gls{lm} approach emerges as the most robust, offering a mechanism that preserves the integrity of the scoring objective while attenuating the influence of unreliable predictions on policy learning. Our findings emphasize that uncertainty is not merely a diagnostic tool but can serve as an active component of the optimization process, improving both the quality and trustworthiness of \textit{de novo} molecular design. We anticipate that these strategies will be broadly applicable to any generative framework in which \gls{rl} is used to guide molecular optimization, whenever a meaningful estimate of predictive uncertainty can be obtained.

\section{Data availability}
The code used in the study is publicly available from the GitHub
repository:
\url{https://github.com/BorjaMedina/UncertaintyAwareRLforCLM}

\printnoidxglossary[type=\acronymtype]

\begin{multicols}{2}
\bibliographystyle{unsrtnat}
\bibliography{referencesPaper}
\end{multicols}

\clearpage
\clearpage

\renewcommand{\figurename}{Supplementary Figure}
\renewcommand{\tablename}{Supplementary Table}

\setcounter{section}{0}
\setcounter{subsection}{0}
\setcounter{subsubsection}{0}
\setcounter{figure}{0}
\setcounter{table}{0}
\setcounter{equation}{0}

\renewcommand{\thesection}{S\arabic{section}}
\renewcommand{\thesubsection}{S\arabic{section}.\arabic{subsection}}
\renewcommand{\thesubsubsection}{S\arabic{section}.\arabic{subsection}.\arabic{subsubsection}}
\renewcommand{\thefigure}{S\arabic{figure}}
\renewcommand{\thetable}{S\arabic{table}}
\renewcommand{\theequation}{S\arabic{equation}}

\begin{center}
    {\LARGE \bfseries Supplementary Information for\par}
    \vspace{1em}
    {\LARGE Uncertainty-aware reinforcement learning for chemical language models\par}
    \vspace{1.5em}

    {\large
    Borja Medina\par
    \vspace{0.5em}
    \addMolAI\par
    \addAstra\par
    \addAddress\par
    \texttt{borja.medinadelasheras@astrazeneca.com}\par
    }

    \vspace{1em}

    {\large
    Jon Paul Janet\par
    \vspace{0.5em}
    \addMolAI\par
    \addAstra\par
    \addAddress\par
    }
\end{center}

\vspace{2em}

\clearpage
\section{Dataset cleaning} \label{SI_datasetCleaning}

Bioactivity datasets for \gls{drd2} and \gls{egfr} were retrieved from ChEMBL version 34. Datasets were processed using a standardized curation pipeline in Python. For both targets, each entry was filtered to retain those with non-missing values for SMILES, Molecule ChEMBL ID, and pChEMBL Value. Chemical validity was then enforced by removing molecules with invalid SMILES strings (RDKit parsing failure).

To standardize molecular representation, valid SMILES were converted to canonical SMILES and  desalted using RDKit. Duplicate entries were resolved in two sequential steps:
    \begin{enumerate}
        \item Aggregation by Molecule ChEMBL ID, retaining the canonical SMILES and the median pChEMBL Value.
        \item Aggregation by canonical SMILES, retaining the first Molecule ChEMBL ID and the median pChEMBL Value.
    \end{enumerate}
    
With this procedure we make sure we end up with one potency value per unique canonical compound.

For the creation of the ChemProp $\gls{egfr}_{\text{full}}$ , the downloaded \gls{egfr} curated dataset was randomly split into train/validation/test partitions using a fixed random seed (42) with an 80/10/10 ratio. Resulting cleaned datasets were exported for downstream experiments.

To create the $\gls{egfr}_{\text{small}}$  the training subset of the dataset was sub-sampled down to $800$ compounds.

Additionally, the cleaned \gls{egfr} and \gls{drd2} datasets were sub-sampled to $2\,200$ random \gls{smi}s to be used as the training datasets for the \gls{ms}.

\clearpage
\section{Different functions and optimization objectives for the different RL runs}

Each raw property was mapped to the $[0,1]$ interval. This transformation ensures that heterogeneous properties, such as physicochemical descriptors and activity predictions, contribute on a comparable scale. Depending on the scoring function, either a sigmoid or a double sigmoid transformation was applied.

For properties with a preferred range of values, we made use of a double sigmoid transformation, which penalizes deviations on either side. This transformation was applied to LogP, BertzCT, \gls{mw}, and \gls{tpsa}.

For properties where larger values were preferred above a specific threshold, we used a sigmoid transformation. In this work, this transformation was applied to the ChemProp activity score.

\begin{table}[h]
\centering
\caption{Transformation functions and parameters used for each scoring component.}
\label{tab:transformations}
\begin{tabularx}{0.8\linewidth}{llX}
\toprule
Component & Transformation & Parameters \\
\hline
LogP & Double sigmoid & low=2, high=3, coef\_div=40, coef\_si=200, coef\_se=200 \\
BertzCT & Double sigmoid & low=800, high=1000, coef\_div=500, coef\_si=15, coef\_se=15 \\
MW & Double sigmoid & low=400, high=650, coef\_div=1000, coef\_si=20, coef\_se=20 \\
TPSA & Double sigmoid & low=80, high=100, coef\_div=100, coef\_si=25, coef\_se=25 \\
ChemProp activity & Sigmoid & low=3, high=9, k=0.18 \\
\hline
\end{tabularx}
\end{table}

Each \gls{rl} run was performed using REINVENT's staged learning framework with a DAP learning strategy ($\sigma = 120$, learning rate $= 1~\times~10^{-4}$). A batch of 64 molecules was generated per step, making use of the \gls{smi} randomization and discarding duplicate sequences within each batch. Scaffold diversity was encouraged through the use of the Identical Murcko Scaffold diversity filter, with a bucket size of 25 and a penalty multiplier of 0.5 applied to molecules that exceeded the bucket capacity.

\clearpage

\section{ChemProp model training and uncertainty calibration}

ChemProp was used as a regression network trained with the \gls{mve} objective, 
such that the model predicted both the activity value and an associated uncertainty estimate. 
Training was performed using the ChemProp CLI, with \texttt{canonical\_smiles} as the molecular input column, \texttt{pChEMBL Value} as the activity target column, and \texttt{split} as the 
predefined data-split column. A single checkpoint, selected by best validation performance, was retained for all subsequent predictions and calibration. Each model was trained for 200 epochs, with the best-performing checkpoint saved according to validation loss. Unless otherwise specified, default ChemProp 
hyperparameters were used.

The software environment comprised ChemProp 2.2.1, Python 3.11.14, PyTorch 2.9.1+cu128, CUDA runtime 12.8, and RDKit 2025.09.1. Training was executed on an NVIDIA A100 GPU.

Uncertainty calibration was performed post hoc using the test split. The selected checkpoint was loaded in evaluation mode with gradient computation disabled. For each molecule in the 
test split, the model produced an \gls{mve} prediction alongside its associated uncertainty estimate using ChemProp's \texttt{MVEEstimator}. These predictions and uncertainty estimates, 
together with the corresponding test targets, were used to fit a \texttt{MVEWeightingCalibrator}. 
The fitted calibrator was serialized to disk and subsequently reused during inference to produce calibrated uncertainty estimates.

\clearpage

\section{Model System with unique scoring function with a mid-initial-distances landscape}

\begin{figure*}[htbp]
    \centering
    \includegraphics[page=1,width=0.8\linewidth]{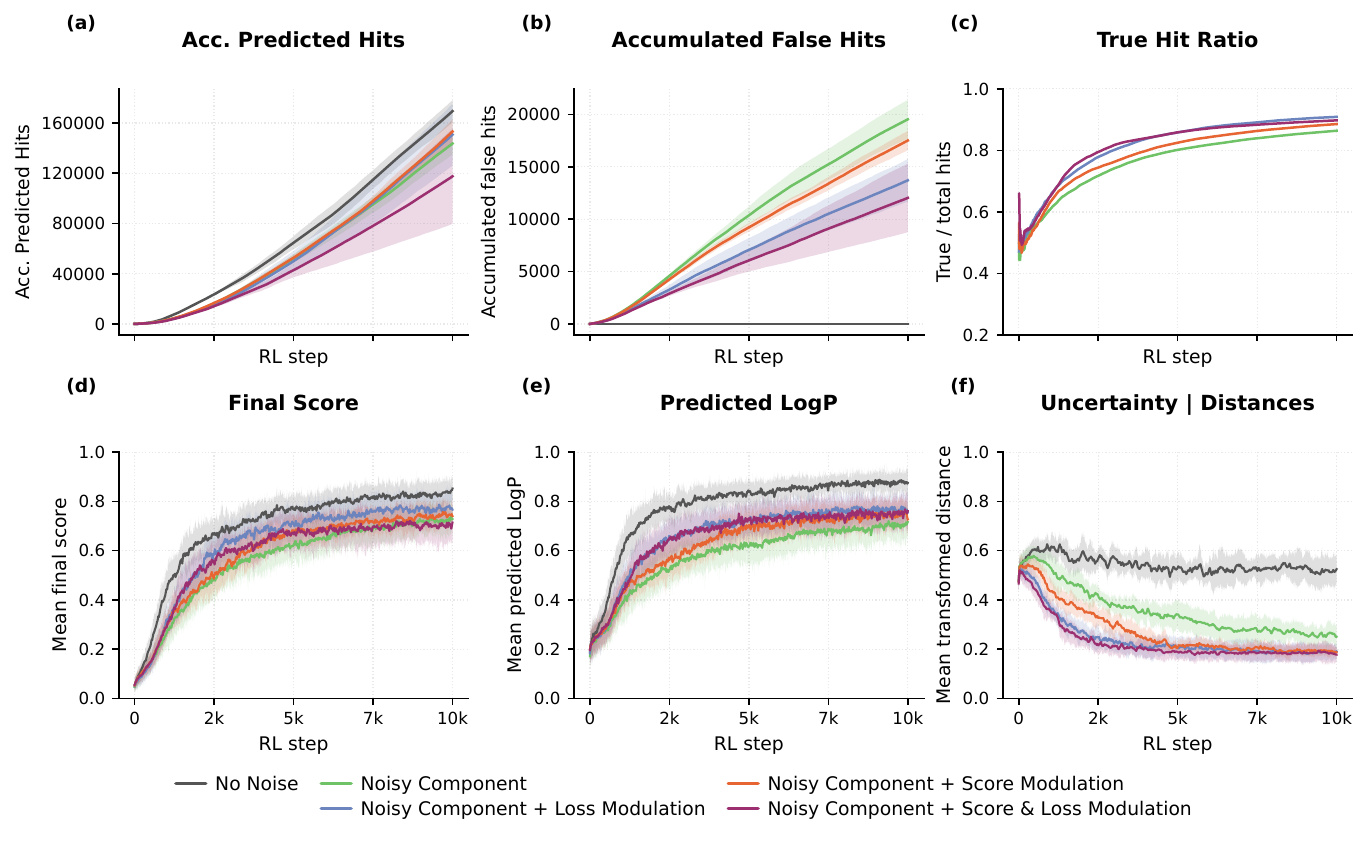}
    \caption{Results for \gls{ms} with one noisy scoring component using RDKit \textit{logP} as the  scoring predictor. We report (a) the total number of accumulated scaffold hits, (b) the number of false hits among them, and (c) the ratio of true to total accumulated hits. Additionally, we report several metrics throughout the \gls{rl} run, including (d) the total RL score guiding the optimization process, (e) the mean transformed predicted \textit{logP} score (including predictive error), and (f) the mean transformed distance, which in \gls{ms} corresponds to the uncertainty measure. For all reported scores, except uncertainty, values closer to 1 indicate better performance. Results are averaged over five independent runs, with the standard deviation shown as the shaded area.}
    \label{fig:resultsHits_Mid_ext}
\end{figure*}
\clearpage


\section{Model System with unique scoring function with a high-initial-distances landscape}

\begin{figure}[htbp]
    \centering
    \includegraphics[page=1,width=0.8\linewidth]{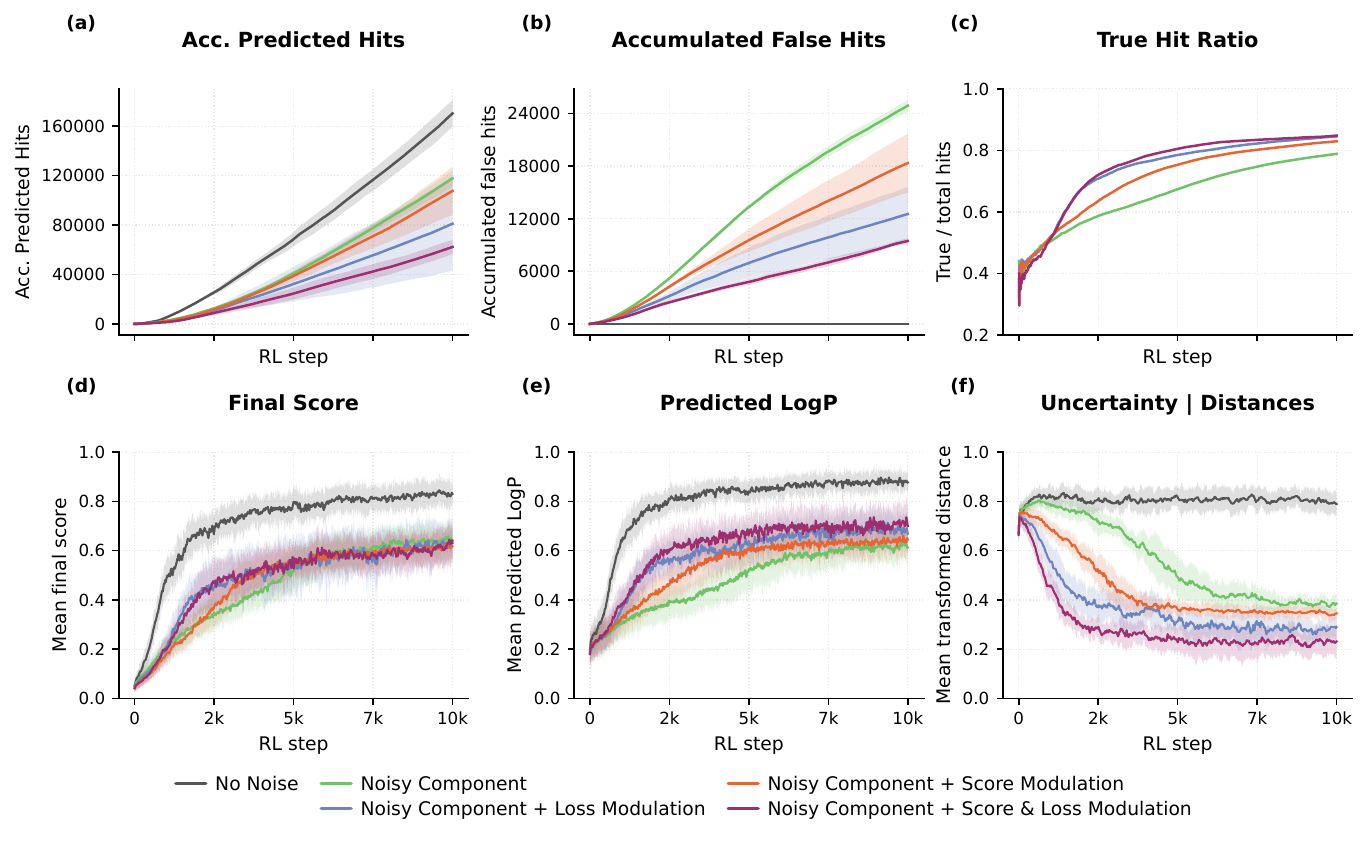}
    \caption{Results for the \gls{ms} setting with one noisy scoring component (\textit{logP} predictor) under a high-initial-distances landscape. In the \gls{ms} setting, high initial distances correspond to high initial uncertainties. We still observe the previously described self-correction behavior, where the noisy component alone is eventually able to reach low-uncertainty regions, although convergence takes longer due to the larger initial shift. The uncertainty-aware methods individually continue to perform well, reaching low-uncertainty regions rapidly while producing fewer false hits. In contrast, the combined method (\gls{sm} + \gls{lm}) performs poorly, yielding substantially fewer hits than the other methods.}
    \label{resultsUniqueNoisy_low}
 \end{figure}
\clearpage


\section{Comparison of distances densities for mid-initial-distances vs high-initial-distances landscapes}

\begin{figure}[htbp]
    \centering
    \includegraphics[page=1,width=0.8\linewidth]{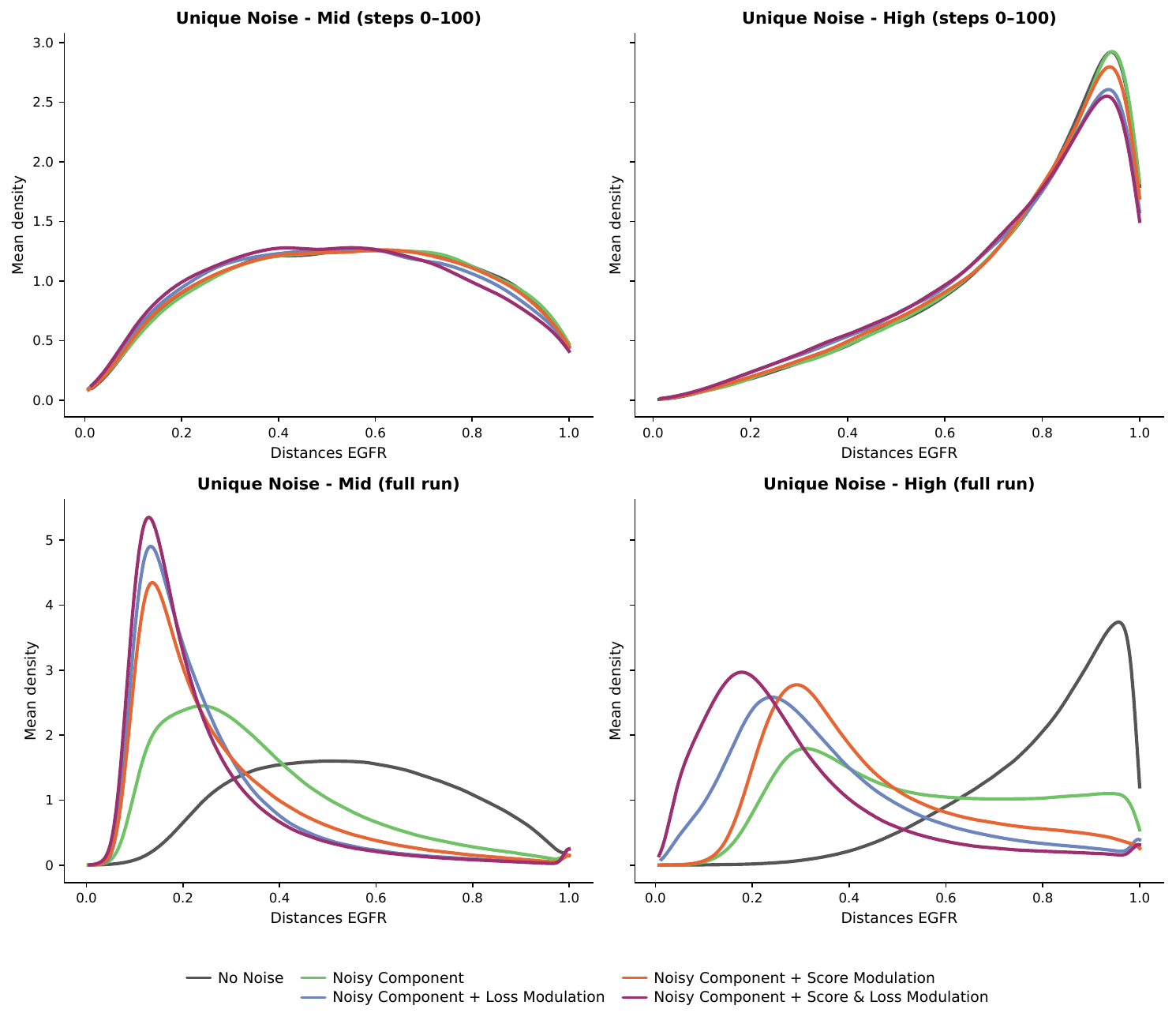}
    \caption{Distribution of distances for the \textit{mid-initial-distances} and \textit{high-initial-distances} settings using the \gls{ms} with one noisy scoring component (\textit{logP} RDKit descriptor). During the first 100 steps, the mid-initial-distances setting produces distances spread across the full $[0,1]$ range, while the high-initial-distances setting concentrates them around a higher mean (0.9). After the full run (10,000 steps), the no-noise baseline keeps the distribution seen at the first 100 steps, in the mid-initial-distances shows distances distributed broadly  but in the high-initial-distances case they are concentrated at very high values. Instead, adding the noisy scoring component alone shifts the density toward lower-uncertainty regions; however, this effect is substantially weaker in the high-initial-distances setting. In contrast, all uncertainty-guided strategies successfully redirect exploration toward low-uncertainty regions in both cases.}
    \label{densities_similaritiesUniqueNoisy}
\end{figure}

The sigmoid transformation in \textit{Eq.~11} is implemented via a reparameterization that maps three hyperparameters $(\mathit{high},\, \mathit{low},\, k)$ onto the standard logistic form:

\begin{equation}
    \beta = \frac{\mathit{high} + \mathit{low}}{2},
    \qquad
    \alpha = \frac{10\,k\,\ln 10}{\mathit{high} - \mathit{low}}.
\end{equation}

We use two different configurations:

\paragraph{Configuration 1 --- mid-initial-distances}
$(\mathit{high},\,\mathit{low},\,k) = (1.5,\;0.6,\;0.3)$:
\[
    \beta_1 = \frac{1.5+0.6}{2} = 1.05,
    \qquad
    \alpha_1 = \frac{10\times0.3\times\ln 10}{1.5-0.6} \approx 7.68.
\]
The score equals $0.5$ when $d=1.05$.

\paragraph{Configuration 2 --- high-initial-distances}
$(\mathit{high},\,\mathit{low},\,k) = (1.3,\;0.5,\;0.3)$:
\[
    \beta_2 = \frac{1.3+0.5}{2} = 0.90,
    \qquad
    \alpha_2 = \frac{10\times0.3\times\ln 10}{1.3-0.5} \approx 8.64.
\]
Here the inflection occurs at $d=0.90$, corresponding to a tighter applicability domain; additionally, the slightly larger $\alpha_2$ produces a sharper transition.

Small changes in this hyperparmeters produce drastic changes in the initial distribution of distances. 

\clearpage
\section{Model System with multiple noisy scoring function with a mid-initial-distances landscape - Extended figure}

\begin{figure}[htbp]
    \centering
    \includegraphics[page=1,width=0.8\linewidth]{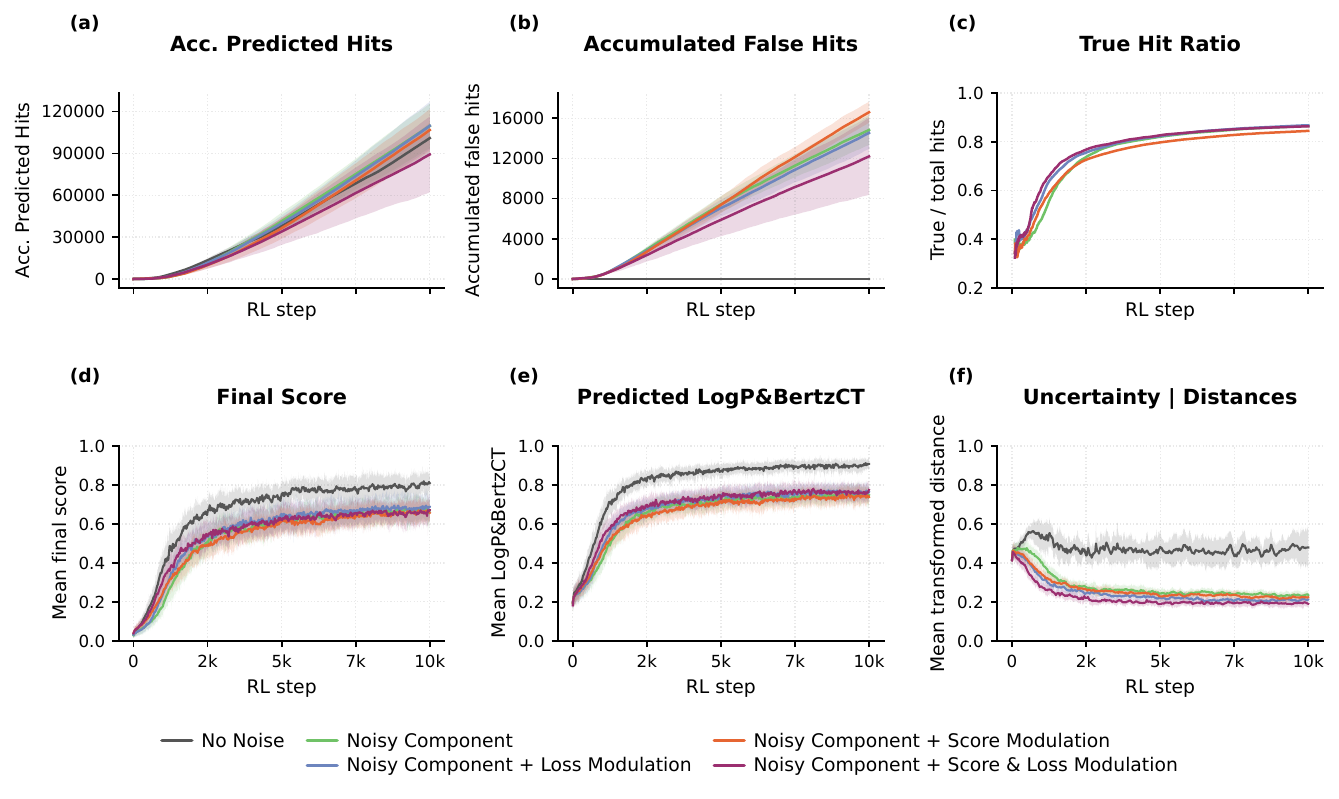}
    \caption{Results for the \gls{ms} setting with two noisy scoring components (\textit{logP} and \textit{BertzCT} RDKit descriptors)  under a mid-initial-distances landscape. We observe the self-correction behavior, where the noisy component alone is eventually able to reach low-uncertainty regions. The number of epochs of difference between uncertainty aware strategies and the Predictive scorinf function on his own is not so big, which consequently forces less differences in the rest of the plots, because we reach similar behaviour.}
    \label{resultsMultiNoisy_midFull}
\end{figure}
\clearpage

\section{Model System with multiple noisy scoring function with a high-initial-distances landscape}

\begin{figure}[htbp]
    \centering
    \includegraphics[page=1,width=0.8\linewidth]{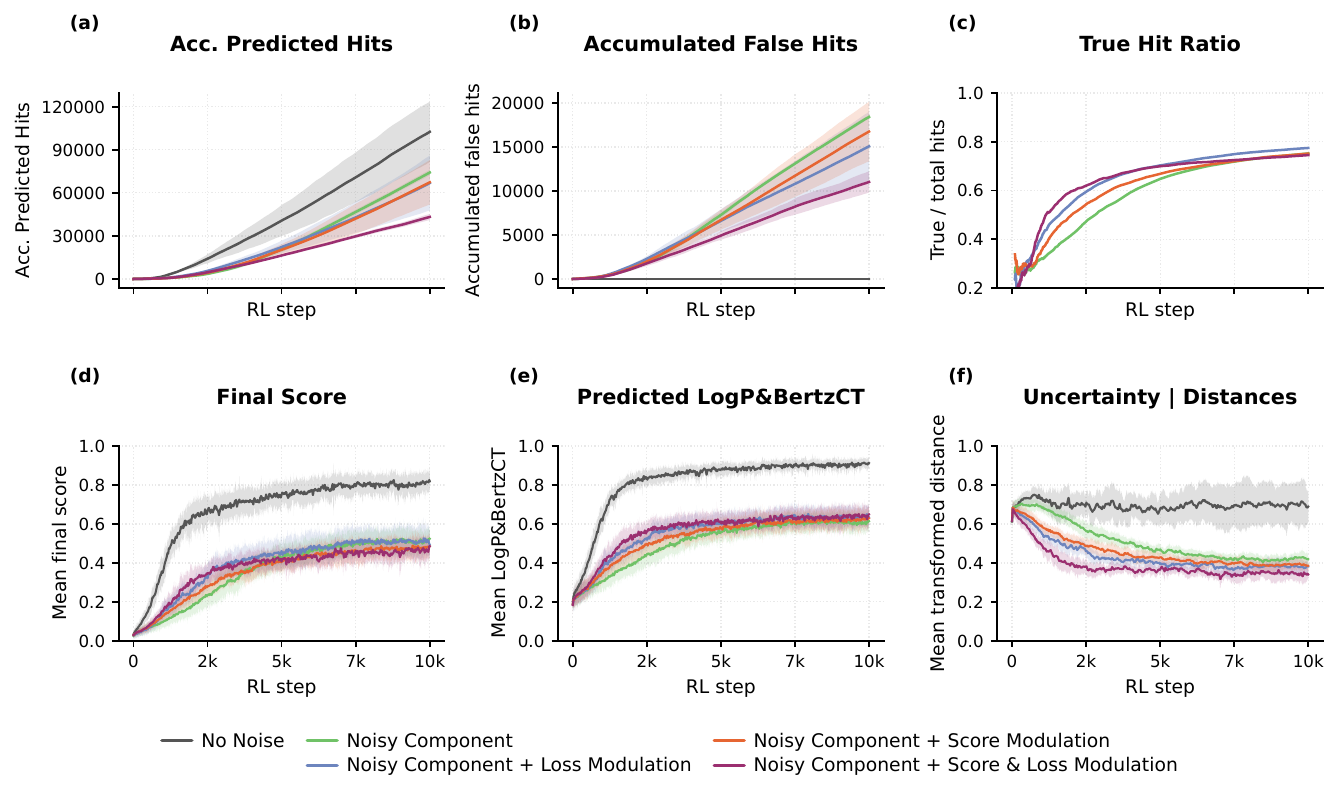}
    \caption{Results for the \gls{ms} setting with two noisy scoring components (\textit{logP} and \textit{BertzCT} RDKit descriptors)  under a high-initial-distances landscape. In the \gls{ms} setting, high initial distances correspond to high initial uncertainty. We still observe the previously described self-correction behavior, where the noisy component alone is eventually able to reach low-uncertainty regions, although convergence takes longer due to the larger initial shift. The uncertainty-aware methods individually continue to perform well, reaching low-uncertainty regions rapidly while producing fewer false hits. In contrast, again, the combined method (\gls{sm} + \gls{lm}) performs poorly, yielding substantially fewer hits than the other methods.}
    \label{resultsMultiNoisy_low}
\end{figure}
\clearpage


\section{Interpreting probabilistic scoring components}

Using probabilistic scoring functions is the mathematically consistent approach when a scoring function returns a full predictive distribution rather than a single point estimate. In contrast, the default implementation in \textit{REINVENT4} applies the score transformation only to a point estimate, which provides a practical approximation but does not formally propagate predictive uncertainty through the \gls{mpo} calculation.

In our case, we therefore implemented a probabilistic scoring function formulation that propagates the full predictive distribution through the scoring pipeline and estimates the final \gls{mpo} score as the mean over sampled realizations, as described in \textit{Section~2.2.1}.

Here, we compare this probabilistic formulation with the default point-estimate setting used in \textit{REINVENT4}. Across our experiments, we did not observe substantial differences between the two approaches in the overall optimization behavior or in the main trends of the evaluated metrics. This indicates that, in the present setting, the point-estimate approximation has only a limited practical impact, even though the probabilistic formulation remains as the correct formulation.

The corresponding results are shown in the following figures.

\clearpage

\subsection{Results for Probabilistic scoring function without uncertainty guidance}
\begin{figure}[htbp]
    \centering
    \includegraphics[page=1,width=0.8\linewidth,trim=0cm 0cm 0cm 0cm,clip]{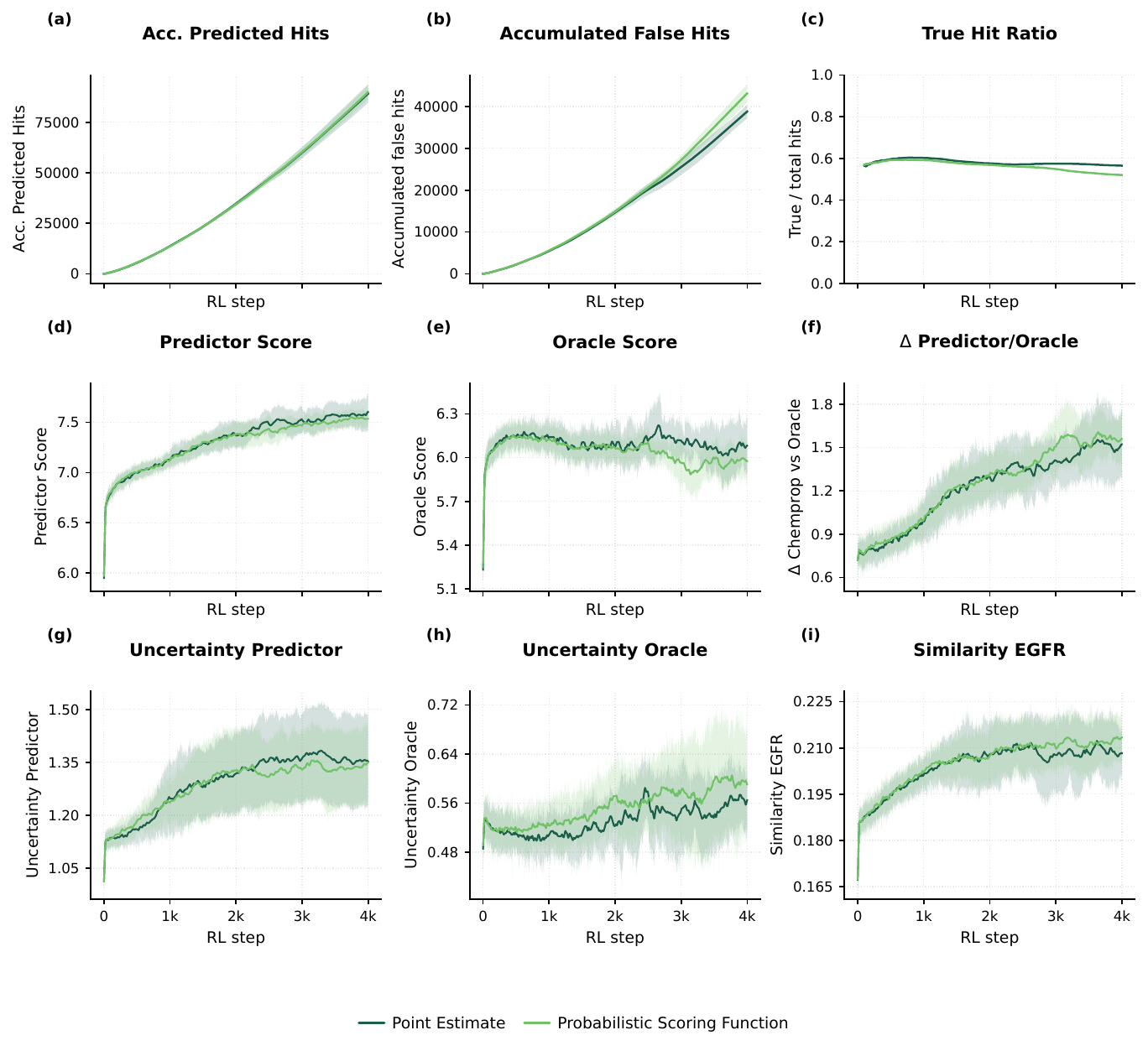}
    \caption{Results for Point vs Probabilistic estimation of the MPO for a scoring function without the uncertainty guidance ("Noisy Component"). In \textit{dark-green} the Point estimation. In \textit{light-green} the probabilistic scoring function.}
    \label{fig:resultsAvNN}
\end{figure}

\clearpage
\subsection{Results for Probabilistic scoring function with the Score modulation strategy}
\begin{figure}[htbp]
    \centering
    \includegraphics[page=1,width=0.8\linewidth,trim=0cm 0cm 0cm 0cm,clip]{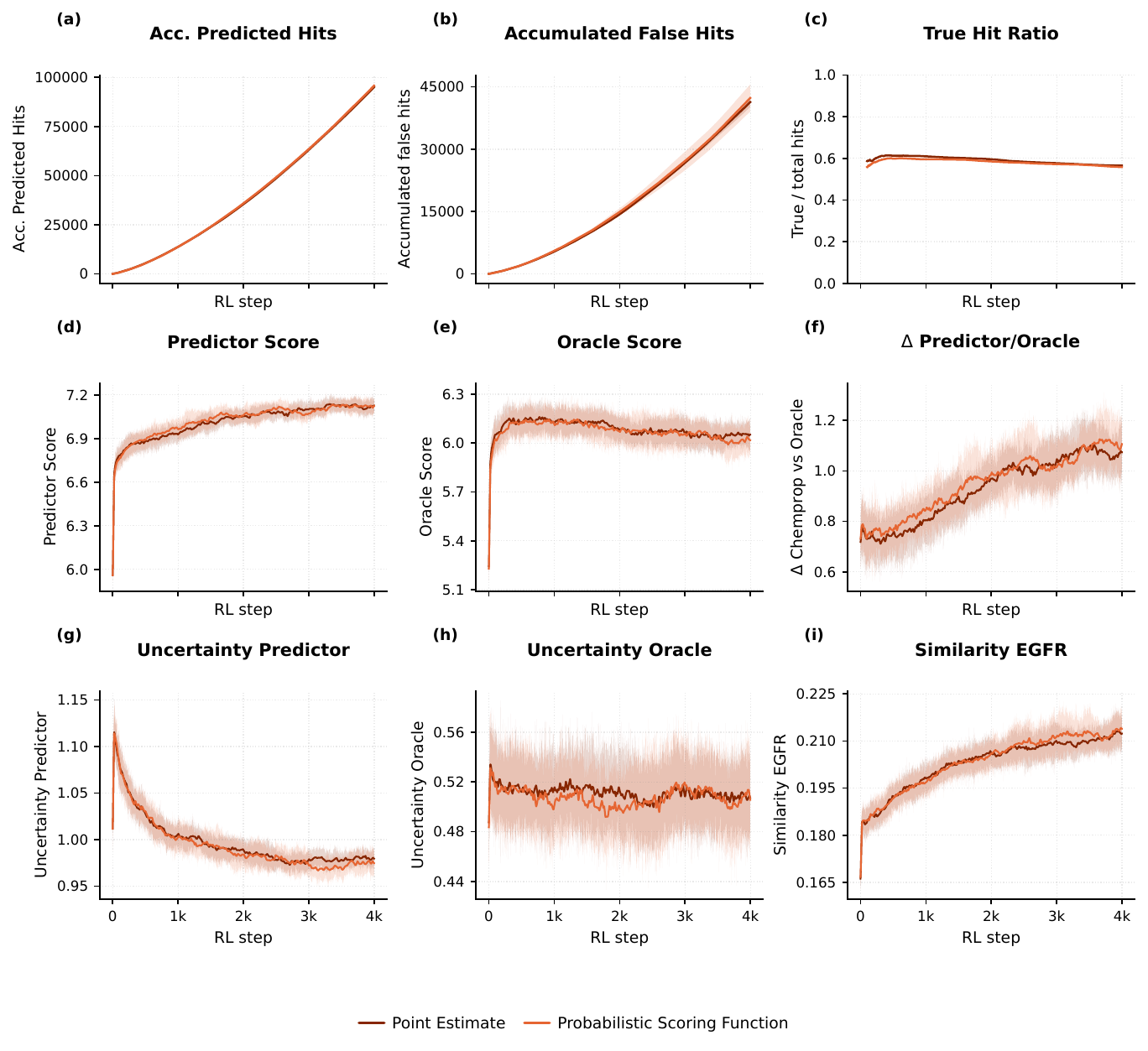}
    \caption{Results for Point vs Probabilistic estimation of the MPO for a scoring function with the Score Modulation strategy. In \textit{dark-orange} the Point estimation. In \textit{light-orange} the probabilistic scoring function.}
    \label{fig:resultsAvSM}
\end{figure}

\clearpage
\subsection{Results for Probabilistic scoring function with the Loss modulation strategy}
\begin{figure}[htbp]
    \centering
    \includegraphics[page=1,width=0.8\linewidth,trim=0cm 0cm 0cm 0cm,clip]{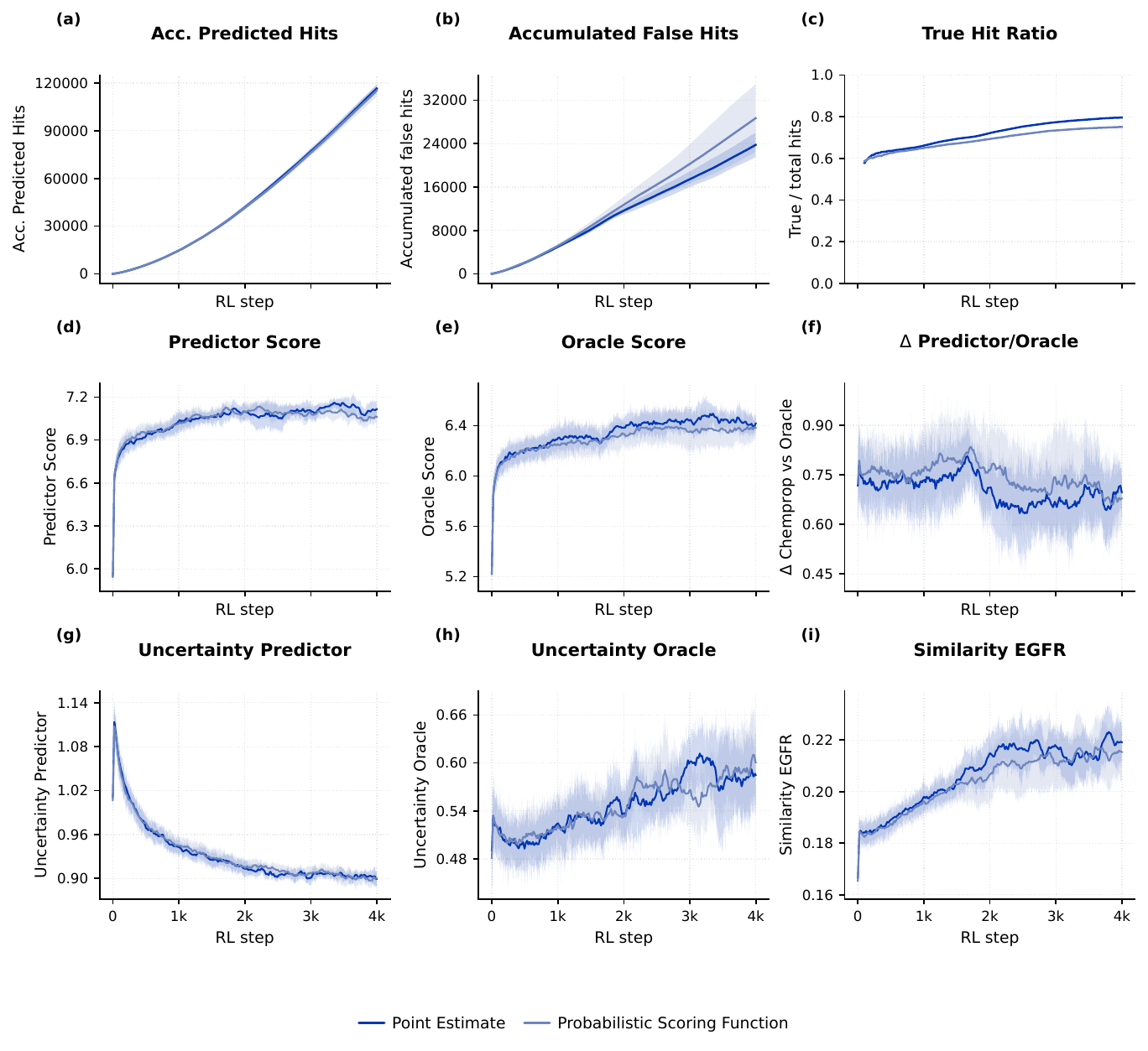}
    \caption{Results for Point vs Probabilistic estimation of the MPO for a scoring function with the Loss Modulation strategy. In \textit{dark-blue} the Point estimation. In \textit{light-blue} the probabilistic scoring function.}
    \label{fig:resultsAvLM}
\end{figure}
\clearpage

\subsection{Results for Probabilistic scoring function with the Score Modulation + Loss modulation strategy}
\begin{figure}[htbp]
    \centering
    \includegraphics[page=1,width=0.8\linewidth,trim=0cm 0cm 0cm 0cm,clip]{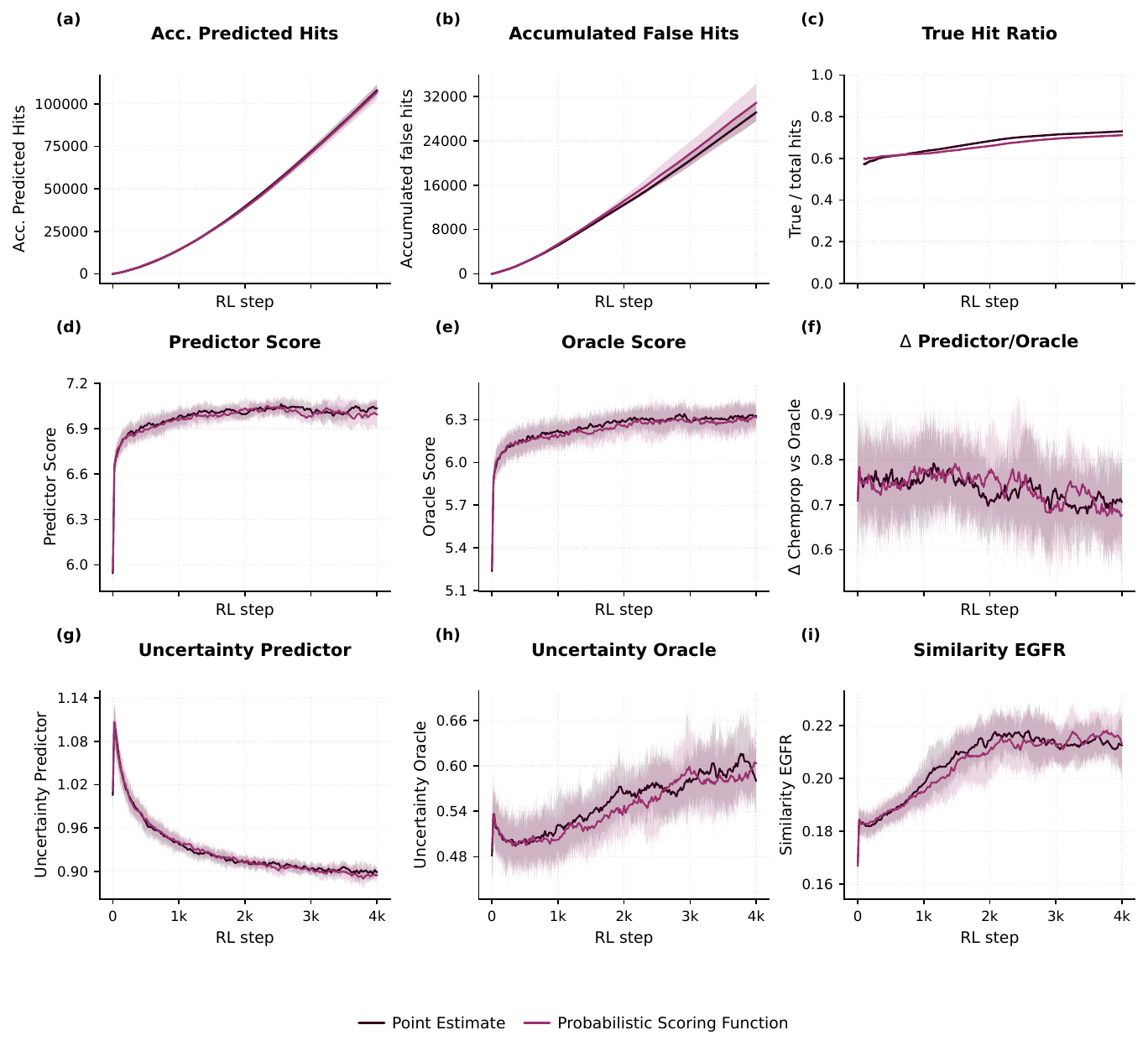}
    \caption{Results for Point vs Probabilistic estimation of the MPO for a scoring function with the Score Modulation + Loss Modulation strategy. In \textit{dark-plum} the Point estimation. In \textit{light-plum} the probabilistic scoring function.}
    \label{fig:resultsAvBoth}
\end{figure}

\end{document}